\documentclass[10pt,journal,compsoc]{IEEEtran}

\ifCLASSOPTIONcompsoc
  \usepackage[nocompress]{cite}
\else
  \usepackage{cite}
\fi

\ifCLASSINFOpdf

\else

\fi

\usepackage{epsfig}
\usepackage{graphicx}
\usepackage{amsmath}
\usepackage{amssymb}
\usepackage{booktabs}
\usepackage{tabularx}
\usepackage{comment}
\usepackage{enumitem}
\usepackage{multirow}
\usepackage{booktabs}
\usepackage{array, caption, threeparttable}
\usepackage{caption}
\usepackage[ruled]{algorithm2e}
\usepackage[table]{xcolor} 

\usepackage{listings}
\usepackage{color}
\usepackage{bbding}
\usepackage{mathtools}
\usepackage{pifont}
\usepackage{todonotes}
\usepackage{gensymb}
\usepackage{coloredtheorem}
\usepackage{hyperref}
\hypersetup{
    colorlinks=true,
    linkcolor=blue,
    filecolor=magenta,      
    urlcolor=cyan,
}
\usepackage{amsmath}
\usepackage{amsthm}
\usepackage{amsfonts}
\newtheorem{definition}{Definition}
\newtheorem{assumption}{Assumption}
\newtheorem{lemma}{Lemma}
\newtheorem{theorem}{Theorem}
\newtheorem{proposition}{Proposition}

\newtheorem{remark}{Remark}
\newtheorem{recall}{Recall}

\definecolor{lightblue}{RGB}{220,240,255}
                                 
\begin{document}

\title{Learning Representation and Synergy Invariances: 
A Povable Framework for Generalized Multimodal Face Anti-Spoofing}

\author{Xun Lin, Shuai Wang, Yi Yu, Zitong Yu, Jiale Zhou, Yizhong Liu, \\Xiaochun Cao, Alex Kot,~\IEEEmembership{Life Fellow,~IEEE}, Yefeng Zheng,~\IEEEmembership{Fellow,~IEEE}

\thanks{
This work was conducted while Xun Lin was a visiting scholar at Westlake University, Hangzhou, China (hosted by Prof. Yefeng Zheng).

The corresponding author is Zitong Yu (email: zitong.yu@ieee.org).
}

\IEEEcompsocitemizethanks{

\IEEEcompsocthanksitem 
Xun Lin, Shuai Wang, and Yizhong Liu are with Beihang University, Beijing, China. (emails: \{linxun, wangshuai\}@buaa.edu.cn). 

\IEEEcompsocthanksitem 
Yi Yu and Alex Kot are with the Rapid-Rich Object Search (ROSE) Lab, Nanyang Technological University, Singapore.

\IEEEcompsocthanksitem 
Zitong Yu is with the Great Bay University, Dongguan, China.

\IEEEcompsocthanksitem 
Jiale Zhou and Yefeng Zheng are with Westlake University, Hangzhou, China.

\IEEEcompsocthanksitem 
Xiaochun Cao is with the Shenzhen Campus of Sun Yat-sen University, Shenzhen, China.

}
}

\markboth{Under Review}%
{Shell \MakeLowercase{\textit{et al.}}: Bare Advanced Demo of IEEEtran.cls for IEEE Computer Society Journals}

\IEEEtitleabstractindextext{%
\begin{abstract}

Multimodal Face Anti-Spoofing (FAS) methods, which integrate multiple visual modalities, often suffer even more severe performance degradation than unimodal FAS when deployed in unseen domains. 
This is mainly due to two overlooked risks that affect cross-domain multimodal generalization.
The first is the \textit{modal representation invariant risk}, i.e., whether representations remain generalizable under domain shift. 
We theoretically show that the inherent class asymmetry in FAS (diverse spoofs vs. compact reals) enlarges the upper bound of generalization error, and this effect is further amplified in multimodal settings.
The second is the \textit{modal synergy invariant risk}, where models overfit to domain-specific inter-modal correlations. 
Such spurious synergy cannot generalize to unseen attacks in target domains, leading to performance drops.
To solve these issues, we propose a provable framework, namely Multimodal \textbf{R}epresentat\textbf{i}on and \textbf{S}ynergy Invariance L\textbf{e}arning (RiSe).
For representation risk, RiSe introduces Asymmetric Invariant Risk Minimization (AsyIRM), which learns an invariant spherical decision boundary in radial space to fit asymmetric distributions, while preserving domain cues in angular space. 
For synergy risk, RiSe employs Multimodal Synergy Disentanglement (MMSD), a self-supervised task enhancing intrinsic, generalizable modal features via cross-sample mixing and disentanglement. 
Theoretical analysis and experiments verify RiSe, which achieves state-of-the-art cross-domain performance. 

\end{abstract}

}

\maketitle

\IEEEdisplaynontitleabstractindextext

%
\IEEEpeerreviewmaketitle

\ifCLASSOPTIONcompsoc
\IEEEraisesectionheading{\section{Introduction}\label{sec:introduction}}
\else
\section{Introduction}
\fi

Face recognition (FR) is primarily used for identity authentication~\cite{handbook}. 
Due to its convenience and accuracy, FR systems have been widely applied in scenarios such as surveillance, mobile payments, and access control~\cite{rizhao2025tpami}. 
In these scenarios, identity theft can lead to severe consequences, such as unauthorized payments and illegal intrusions, making the security of FR systems critically important. 
However, early studies have shown that FR systems are vulnerable to face presentation attacks (PAs), including printed photos, video replay, and 3D wearable masks~\cite{survey}. 
Such vulnerabilities pose a security threat to various industries, including finance, transportation, and security~\cite{handbook}. 
To address this issue, Face Anti-Spoofing (FAS) techniques have been developed to protect FR systems from PAs~\cite{survey}.
\begin{figure*}[t]
    \centering
    \includegraphics[width=\linewidth]{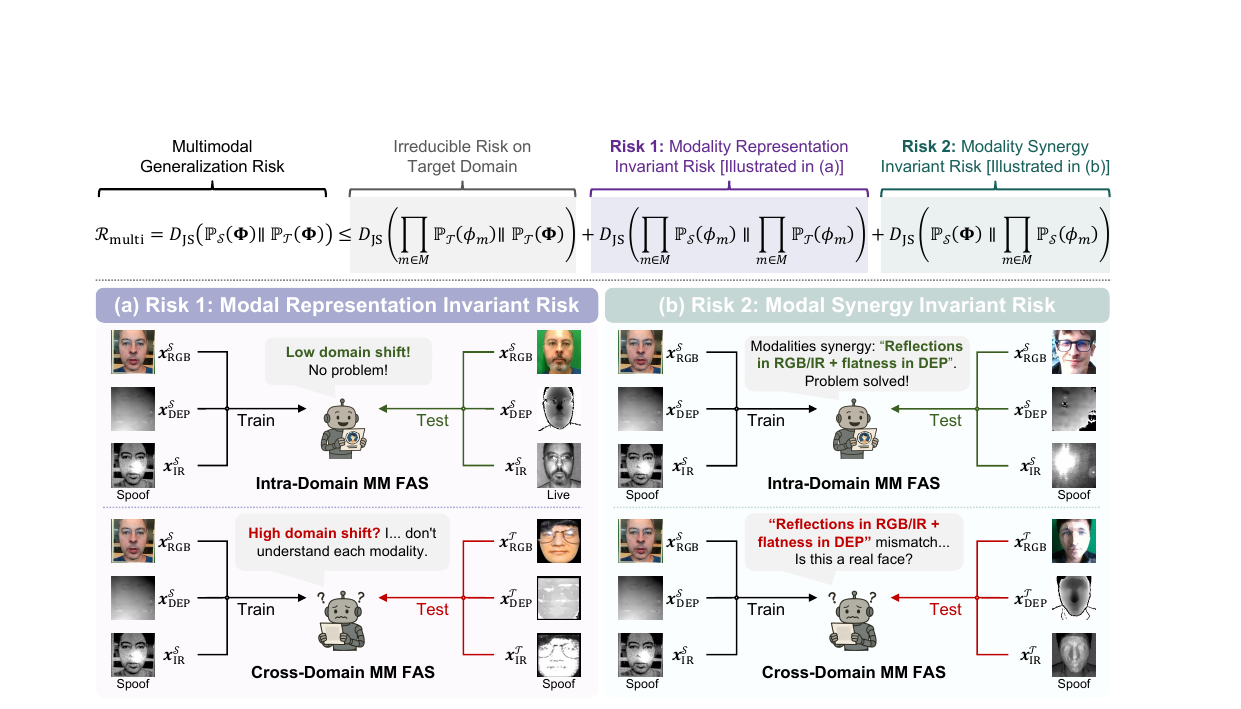}
    \caption{
    Illustration of our decomposition of the multimodal generalization risk into two trainable invariant risks. 
    (a) The modal representation invariant risk \textit{(Risk 1)} arises when unimodal representations learned on a source domain ($\mathcal S$) fail to generalize to a target domain ($\mathcal T$) due to a large domain shift. 
    (b) The modal synergy invariant risk \textit{(Risk 2)} occurs when a spurious cross-modal correlation (synergy) learned in $\mathcal S$ proves invalid in $\mathcal T$, leading to shortcut-based prediction errors.
 }
    \label{fig:main}
\end{figure*}

Early handcrafted-feature-based FAS methods demonstrated limited representational capacity, making it difficult for their detection performance to meet practical requirements~\cite{texture,s-adapter}.
With the successful application of deep learning in computer vision~\cite{cv-survey}, numerous data-driven deep-learning-based unimodal FAS methods (relying solely on visible light images) have been proposed~\cite{deep-pixel,mc-face-pad}, achieving inspiring progress.
Although these methods perform well in intra-domain scenarios, where deployment environments and attack types are known~\cite{survey,dg2019}, their generalization capability remains insufficient in cross-domain scenarios, such as when deployed in environments different from the training setting or confronted with unseen attacks~\cite{one-class}.
More recently, many FAS methods have introduced domain generalization (DG) techniques to improve cross-domain robustness, e.g., by integrating adversarial training~\cite{dg2019,conditional-feature,ssdg} or feature disentanglement~\cite{ssan,cfpl,trace-distanglement} to learn domain-invariant representations. 
However, these methods still tend to produce incorrect detection results when faced with samples exhibiting severe domain shifts.

The growing sophistication of PAs~\cite{crzfas4} has motivated a shift towards multimodal FAS systems that leverage complementary data modalities, such as RGB, depth, and infrared (IR) imagery~\cite{wmca, cefa}. 
By capturing richer physical cues, i.e., texture from RGB, 3D structure from depth, and thermal patterns from IR, state-of-the-art (SoTA) multimodal approaches have significantly improved intra-domain accuracy~\cite{fm-vit,hd-aaai23}. 
This raises a critical question: \textit{Can the addition of modalities mitigate the persistent cross-domain problem?} 
Unfortunately, previous findings~\cite{mmdg,xun2025tpami} suggest the contrary: multi-modality often introduces greater challenges in cross-domain scenarios, with some multimodal models even underperforming their single-modal counterparts. 
We posit that this paradox stems from two important risks that have been largely overlooked by existing FAS methods.

\textit{Risk 1: Modal Representation Invariant Risk.} 
Data-driven FAS models are highly prone to learning spurious shortcuts from source domains~\cite{dg2019}. 
For instance, as shown in Fig.~\ref{fig:main}(a), a model trained on bright, high-resolution live faces and blurry, low-resolution print attacks might erroneously equate ``high resolution'' with authenticity. 
Such a decision logic, based on domain-specific attributes, fails when confronted with unseen domains. 
To mitigate this risk, recent studies~\cite{safas, dadm} build on Invariant Risk Minimization (IRM)~\cite{irm}, seeking to learn representations and decision boundaries that remain invariant across domains.
However, this paradigm faces a unique challenge in the FAS task, rooted in an inherent class asymmetry: the distribution of presentation attacks, encompassing diverse materials and schemes, is far more extensive and scattered than the compact distribution of live faces~\cite{ssdg,mmdg,xun2025tpami}. 
In Sec.~\ref{sec:method}, we theoretically show that this asymmetry enlarges the upper bound of generalization error, and in the multimodal setting, the effect is further amplified. 
On one hand, combining multiple modalities expands the already vast distribution space of PAs. 
On the other hand, different modalities have unequal sensitivities to domain shifts—for instance, depth and IR are much more vulnerable to illumination and sensor variation than RGB~\cite{xun2025tpami}. 
This produces asymmetric feature distributions that are unevenly distorted across modalities, making symmetric decision boundaries (e.g., linear hyperplanes~\cite{dadm,safas}) ineffective for generalization.

\textit{Risk 2: Modal Synergy Invariant Risk.} 
This risk arises not from the features of any single modality, but from the model overfitting to spurious inter-modal synergetical correlations that are idiosyncratic to the source domains. 
For example, Fig.~\ref{fig:main}(b), a model may learn to classify a face as a spoof by relying on the joint correlation of reflections in RGB/infrared and flatness in depth. 
While individually reliable, such statistical co-occurrence can be coincidental and domain-specific. 
When deployed in a new domain, an unseen attack (e.g., 3D mask attack) may mismatch the spurious correlation, causing the learned synergy to collapse
This causes the previously learned synergistic relationship to ``collapse,'' leading to misclassification, even if each individual modality still carries valid cues. 
Current DG-based FAS methods, which primarily focus on learning invariant representations for a single modality, fail to address the generalization of the synergy itself.

To overcome these two risks, we propose a provable multimodal FAS framework, namely multimodal \textbf{\underline{R}}epresentat\textbf{\underline{i}}on and \textbf{\underline{S}}ynergy Invarianc\textbf{\underline{e}} Learning (\textbf{RiSe}).
First, to minimize \textit{Risk 1}, RiSe introduces an Asymmetric Invariant Risk Minimization (AsyIRM). 
AsyIRM projects the joint multimodal features into a spherical space where all domains share an invariant radius as the decision boundary—features with a norm inside the sphere are classified as genuine, and those outside as PAs. 
This design naturally accommodates the asymmetric distribution of compact live faces versus scattered PAs, a characteristic that is proven to be amplified in the multimodal context. 
Concurrently, we leverage the angular space, which is orthogonal to the radial direction, to preserve and separate domain-specific information. 
This ensures the model learns an invariant radial classification boundary without discarding domain diversity. 
Furthermore, we theoretically prove that this asymmetric design achieves a tighter generalization error bound compared to traditional IRM.

Second, to mitigate \textit{Risk 2}, RiSe incorporates a Multimodal Synergy Disentanglement (MMSD) auxiliary task. 
During training, we randomly mix frequency components from different modalities across different samples and shuffle their order, compelling lightweight decoders to restore their original modal identities and sequence. 
MMSD forces the backbone encoder to learn intrinsic, context-free features of each modality, rather than their spurious inter-modal correlations, thereby acquiring a robust and generalizable synergistic understanding. 
The efficacy of MMSD in mitigating the \textit{Risk 2} is also theoretically justified.

Our contributions can be summarized as follows:
\begin{itemize}
    \item We propose a novel framework, RiSe, to address two overlooked challenges in multimodal cross-domain FAS: the modal representation invariant risk \textit{(Risk 1)} and the modal synergy invariant risk \textit{(Risk 2)}.  

    \vspace{1mm}
    \item We prove that prior IRM-based methods inevitably suffer from an enlarged generalization error bound in FAS due to the inherent class asymmetry between compact live faces and diverse spoof attacks. We also found that this effect is further amplified in multimodal scenarios by uneven domain shifts across modalities.  

    \vspace{1mm}
    \item To mitigate \textit{Risk 1}, we introduce Asymmetric Invariant Risk Minimization (AsyIRM), which learns an invariant radial classifier in spherical space to accommodate asymmetric distributions, while preserving domain-specific information in the angular dimension.  

    \vspace{1mm}
    \item To mitigate \textit{Risk 2}, we design Multimodal Synergy Disentanglement (MMSD), a self-supervised auxiliary task that enforces disentanglement of intrinsic modal features from spurious inter-modal correlations, thus ensuring more generalizable synergy.  

    \vspace{1mm}
    \item We theoretically derive the upper bound of multimodal cross-domain risks, and prove that AsyIRM and MMSD respectively optimize \textit{Risk 1} and \textit{Risk 2} within it.  

    \vspace{1mm}
    \item Extensive experiments on four multimodal FAS DG benchmarks (with four protocols) demonstrate that RiSe achieves SoTA cross-domain generalization.  
\end{itemize}

The remainder of this paper is organized as follows. Section~\ref{sec:relatedwork} provides a survey of related literature. 
In Section~\ref{sec:method}, we present preliminaries and our proposed RiSe framework in detail; this includes a thorough introduction to its core components, AsyIRM and MMSD, accompanied by theoretical analysis. 
Section~\ref{sec:experiment} introduces the benchmark for generalized multimodal FAS and reports extensive experimental results, including comparisons against SoTA methods,  fine-grained ablation studies, and analysis on important hyper-parameters. 
Finally, Section~\ref{sec:conclusion} concludes the paper.
The detailed derivations and proofs are provided in Appendix.

\section{Related Work} \label{sec:relatedwork}
\subsection{Domain Generalized Unimodal Face Anti-Spoofing}
Domain generalization in FAS aims to train a model on multiple source domains that can generalize effectively to unseen target domains~\cite{adaptive-transformer, dg2019}. 
Early efforts in this direction primarily focused on learning a shared, domain-invariant feature space directly from visual data. 
These methods employed various strategies, including adversarial training to confuse a domain discriminator~\cite{trace-distanglement, conditional-feature, cyclically}, meta-learning to simulate domain shifts during training~\cite{meta-teacher, dual-meta, meta-pattern, expert, ebdg}, and disentangling style and content features to learn domain-invariant representations~\cite{ssan, iadg, uda, cfpl}. 
Instead of learning strictly domain-invariant representations, SA-FAS~\cite{safas} retained domain-specific information while enforcing domain invariance at the classifier level via Invariant Risk Minimization (IRM)~\cite{irm}.
To further improve the performance on target domains, test-time adaptation~\cite{tta-fas, optimal2025cvpr} and generalization~\cite{ttdg} methods have been proposed to optimize the FAS model by leveraging unlabeled test samples.

Besides learning-based approaches, other works enhanced the generalization of FAS from a data perspective, aiming to alleviate the scarcity of large-scale face datasets caused by privacy concerns. 
For instance, \cite{rizhao2025ijcv} exploited physical priors of presentation attacks for data augmentation, DiffFAS \cite{difffas} leveraged diffusion models to synthesize high-fidelity and diverse attack samples that better cover cross-domain variations, while AG-FAS \cite{defake} trained a ``de-fake'' generator solely on real faces to detect spoofs as deviations from the learned liveness distribution.

More recently, the paradigm has shifted towards leveraging the powerful visual priors of large-scale Vision-Language Models (VLMs), such as Contrastive Language-Image Pre-training (CLIP~\cite{clip}). 
An initial challenge is to adapt these massive models to the smaller FAS datasets without overfitting, which has been addressed through Parameter-Efficient Transfer Learning (PETL) techniques~\cite{continual, adaptive-transformer, flip, s-adapter, rizhao2025tpami}. 
Building on this, recent works explored using textual guidance to enhance generalization. 
These methods range from contrastive fine-tuning strategies that align image views with text prompts~\cite{flip}, to more sophisticated prompt learning techniques that generate adaptive style and content prompts to guide the model's focus~\cite{cfpl,vlfas}. 
SLIP~\cite{slip} further extends this by using language-guided spoof cue estimation and prompt-driven feature disentanglement within a one-class FAS framework. 
In a novel direction, I-FAS~\cite{ajian2025aaai} and FaceShield~\cite{faceshield} reframed the FAS task as an interpretable visual question answering (VQA) problem, utilizing Multimodal Large Language Models (MLLMs) to provide both a decision and a natural language rationale.

However, these methods are designed for and evaluated in the single-modal (RGB) setting. 
They do not explicitly address the unique challenges that arise in multi-modal generalization, where the complex interplay between modalities introduces new risks. 

\subsection{Multimodal Face Anti-Spoofing}
Multimodal Face Anti-Spoofing (FAS) systems aim to enhance PAs detection by integrating complementary information from diverse sensors, typically RGB, depth, and IR. 
The core premise is that different modalities capture distinct physical cues~\cite{handbook}, and while certain attack traces may be imperceptible in one modality, they can often be revealed by leveraging others~\cite{fm-vit, hyperbolic}.
Early research in this area primarily focused on feature fusion strategies, ranging from simple channel-wise concatenation~\cite{mccnn, mc-face-pad} to more complex late fusion of features from separate extraction branches~\cite{mm-cdcn, facebagnet, kcq-fas1, kcq-fas2}. More recent approaches have introduced sophisticated mechanisms such as attention-based fusion~\cite{hd-aaai23, dual-stream}, adaptive cross-modal loss functions~\cite{cmfl}, and cross-modality translation~\cite{cross-translate, lizhi} to better exploit the complementary nature of the data.

While effective in intra-domain settings, the generalization of these methods to unseen domains presents a significant challenge. Our previous works~\cite{mmdg, xun2025tpami} were the first to identify a crucial paradox: contrary to intuition, the introduction of multiple modalities can exacerbate domain shifts compared to single-modal scenarios. 
To solve this problem, Lin et al.~\cite{mmdg, xun2025tpami} tried to rebalance the discriminative power of each modality during training and suppress unreliable modalities during inference. 
This concern for modality robustness also extends to the flexible-modal FAS setting, where models must contend with incomplete modal inputs during training or testing~\cite{fm-benchmark, vp-fas, ama, fm-vit, ma-vit}. 
Following these insights, DADM~\cite{dadm} further advanced the field by extending the IRM theorem to the multimodal context, aiming to learn a domain-invariant classification hyperplane to enhance generalization capability in multimodal FAS.
Despite this progress, existing multimodal methods still fail to recognize and address the two fundamental risks in multimodal cross-domain FAS, and they also lack the necessary theoretical justification. 

\begin{figure*}[t]
    \centering
    \includegraphics[width=0.95\linewidth]{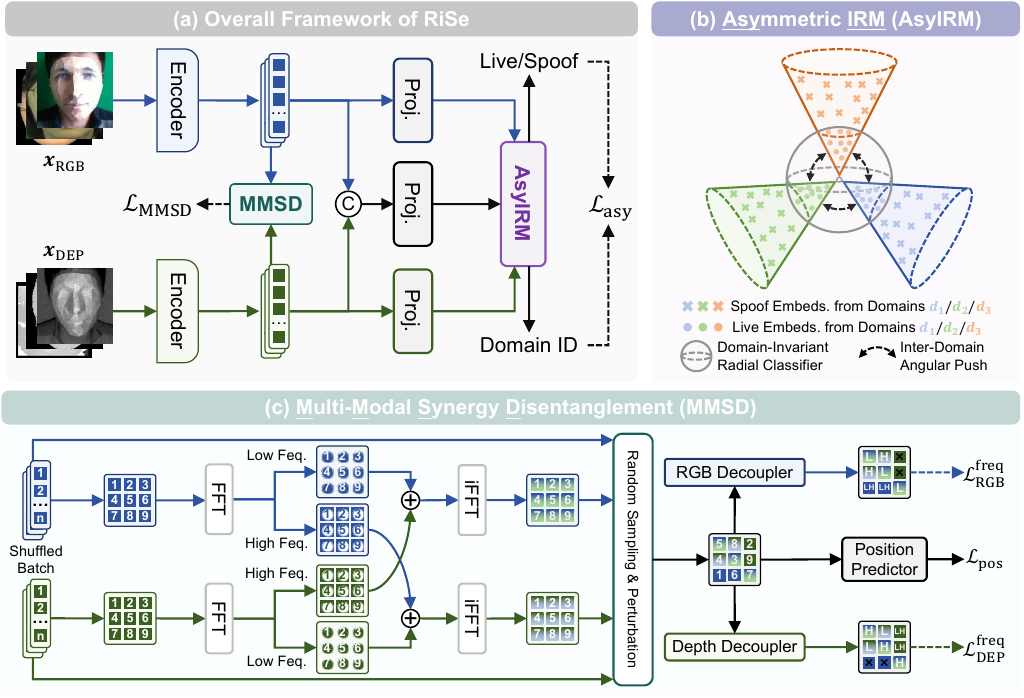}
    \vspace{-0.4em}
    \caption{
    Framework of our proposed RiSe: 
    (a) Overall end-to-end architecture, where modality features are optimized by our two core modules, AsyIRM and MMSD;
    (b) AsyIRM learns a disentangled embedding by using feature norms and domain-invariant radius for the live/spoof classification, and feature directions for a domain-separating angular push;
    (c) MMSD disrupts spurious cross-modal correlations through pretext tasks based on cross-sample frequency mixing and spatial token shuffling.
    For better clarity, we only show the scenario of RGB-Depth FAS here.
    }
    \label{fig:method}
\end{figure*}

\section{Methods} \label{sec:method}
As shown in Fig.~\ref{fig:method}, our RiSe framework is composed of a multi-branch backbone encoder, the proposed AsyIRM (serving as the classifier), and MMSD (as an auxiliary task). 
We begin by introducing the necessary preliminaries in Sec.~\ref{sec:preliminary}, including the formal definition of multimodal cross-domain FAS as well as experience risk minimization (ERM) and IRM. 
Then, in Sec.~\ref{sec:overall}, we present the overall RiSe framework and derive its upper bound of multimodal generalization risks. 
Subsequently, Sec.~\ref{sec:asyirm} and Sec.~\ref{sec:syndis} detail AsyIRM and MMSD, respectively, along with their corresponding theoretical analysis.

\subsection{Preliminary} \label{sec:preliminary}
\subsubsection{Problem Formulation}
Let $\mathcal X_{m} \subset \mathbb R^{H\times W\times C}$ denote the input space of the $m$-th modality, where $m \in \mathcal M\!=\!\{\mathrm{RGB}, \mathrm{DEP}, \mathrm{IR}\}$ corresponding to RGB, depth, and infrared). 
Let $\mathcal Y\!=\!\{0~(\text{live}), 1~(\text{spoof})\}$ be the output space.
A FAS method is given access to a set of training data from $E$ source domains $\mathcal E_{\mathcal S}\!=\!\{e_1, e_2, \cdots, e_{E}\}$, and is evaluated on an unseen target domain $\mathcal E_{\mathcal T}\!=\!\{e^*\}$.
Each sample is represented as $(\boldsymbol x_{\mathrm{RGB}}^i, \boldsymbol x_{\mathrm{DEP}}^i, \boldsymbol x_{\mathrm{IR}}^i, y_i, e_{\varsigma_i})$, where $\boldsymbol x^m_i$ is the observation of the $i$-th sample from modality $m$, $y_i\in\mathcal Y$ is the binary label, and $\varsigma_i\in[1, E]$ is the domain ID, respectively. The training dataset is denoted as:
\begin{equation}
    \mathcal D = \{(\boldsymbol x_{\mathrm{RGB}}^i, \boldsymbol x_{\mathrm{DEP}}^i, \boldsymbol x_{\mathrm{IR}}^i, y_i, e_{\varsigma_i})\}_{i=1}^{N},
\end{equation}
where $N$ is the number of training samples. 
The goal of cross-domain multimodal FAS is to learn a decision function:
\begin{equation}
    f: (\boldsymbol x_{\mathrm{RGB}}^i, \boldsymbol x_{\mathrm{DEP}}^i, \boldsymbol x_{\mathrm{IR}}^i) \rightarrow \mathcal Y,
\end{equation}
which is used to detect whether a multimodal face sample from an unseen domain $e^*$ is live or spoof.

\subsubsection{From ERM to IRM}
We first recall the standard ERM theorem, which is widely applied by existing FAS methods~\cite{mmdg}. For better clarity, we extend the original unimodal version to multimodal:

\begin{definition}[Empirical Risk Minimization, ERM]
\label{definition:erm}
Given a dataset sampled from $E$ source domains 
$\mathcal{E}_{\text{train}}$, 
$\mathcal{D}=\{(\boldsymbol{x}_i, y_i, e_i)\}_{i=1}^N$, 
where $\boldsymbol{x}_i$ denotes a multimodal input 
(e.g., RGB, depth, infrared), 
$y_i \in \mathcal Y$ is the label (live/spoof), and $e_i$ is the domain label.
A decision function $\mathcal{F} = \mathcal B\circ\phi $ is composed of an encoder 
$\phi: \mathcal{X}\rightarrow\mathbb{R}^m$ and a classifier 
$\mathcal B: \mathbb{R}^m\rightarrow \mathcal Y$. 
For each domain $e$, the empirical risk is defined as:
\begin{equation}\label{eq:erm}
    \hat{\mathcal R}_{\mathcal S}(\mathcal{F}) = \frac{1}{E}\sum_{i=1}^{E} 
    \mathcal L\Bigl(\mathcal{F}(\boldsymbol{x}_i), y_i \Bigr).
\end{equation}
\end{definition}

In short, ERM minimizes the average empirical risk over all source domains.
Although ERM provides a simple and effective principle under the i.i.d. setting (intra-domain evaluation), its limitation becomes evident when dealing with distribution shifts across domains. 
By merely minimizing the averaged empirical risk over mixed training data, the model may resort to superficial differences between domains, such as variations in image resolution, illumination, or sensor characteristics, to distinguish live from spoof faces. 
These domain-specific cues are often incidental rather than intrinsic to the spoofing patterns, leading to poor generalization when the model encounters previously unseen domains. 
This observation motivates IRM~\cite{irm,safas,dadm}, which encourages the joint learning of domain-invariant representation and a consistent classifier:

\begin{definition}[Invariant Risk Minimization, IRM]
\label{theorem:irm}
As introduced by Arjovsky et al.~\cite{irm} and adopted in cross-domain FAS~\cite{safas,dadm}, IRM aims to learn a representation $\phi(\cdot)$ such that there exists a global optimal classifier $h$ that performs well across all source domains. 
In vanilla IRM, the invariant classifier is typically a linear hyperplane $\mathcal B_\beta(\boldsymbol z) = \beta^\top \boldsymbol z$, and the objective is formulated as:
\begin{equation}\label{eq:irm}
\begin{aligned}
    \min_{\phi, \beta} \quad 
    & \sum_{e \in \mathcal{E}_{\text{train}}} 
      \hat{\mathcal R}_{\mathcal S}^e(\phi, \mathcal B_\beta), \\
    \text{s.t.} \quad 
    & \beta^* \in 
      \arg\min_{\beta} 
      \hat{\mathcal R}^e(\phi, \mathcal B_{\beta}), \forall e \in \mathcal{E}_{\mathcal S}.
\end{aligned}
\end{equation}
\end{definition}

Different from ERM in Eq.~(\ref{eq:erm}), IRM introduces an additional constraint in order to enforce the invariance of the classification hyperplane across all source domains (see Eq.~(\ref{eq:irm})).
The invariance condition requires that all domain-wise solutions coincide with a global classifier:
\begin{equation}
    \forall e \in \mathcal E_{\mathcal S},~\beta_e = \beta^*.
\end{equation}

This constraint ensures that the representation $\phi(\cdot)$ supports a single classifier that is simultaneously optimal for every source domain. 
Notably, Ahuja et al.~\cite{ahuja2021iclr} proved that when distribution shifts are driven by confounders or anti-causal variables (such as identity, appearance, or sensor characteristics irrelevant to the spoofing label in FAS), IRM can theoretically recover an asymptotically optimal solution that generalizes well to unseen environments. In contrast, ERM may converge to a biased solution under such conditions, which cannot be corrected even with unlimited data.

\subsection{Overall Framework of RiSe} \label{sec:overall}
In this subsection, we first extend the cross-domain generalization risk from the unimodal to the multimodal setting, which clarifies the two fundamental risks that motivate the proposed RiSe. 
We then present an upper bound that decomposes the multimodal risk into interpretable terms, followed by the overview of RiSe shown in Fig.~\ref{fig:method}.

\subsubsection{Generalization Risk: From Unimodal to Multimodal}  

First, we define the generalization error risk as the divergence between the source and target joint representational distributions:

\begin{definition}[Unimodal Generalization Risk]
    \label{def:dg}
    \textit{The unimodal generalization risk $\mathcal R_{\mathrm{uni}}$ is captured by the discrepancy between the source and target distributions of a single modality feature $\phi$: }
    \begin{equation}
        \mathcal R_{\mathrm{uni}} = D_{\textrm{JS}}\bigl(\mathbb P_{\mathcal S}(\phi)\,\|\,\mathbb P_{\mathcal T}(\phi)\bigr),
    \end{equation}
    \textit{where $D_{\text{JS}}(\cdot\|\cdot)$ denotes the Jensen–Shannon (JS) divergence~\cite{js}, $\mathcal S$ represents the source domains, and $\mathcal T$ denotes the target domain.}
\end{definition}

\begin{definition}[Multimodal Generalization Risk]
    \label{def:mmdg}
    \textit{For the multimodal case, let the multimodal joint representation $\boldsymbol \Phi =\{\phi_1, \phi_2, \cdots, \phi_m\}$, the multimodal generalization error risk can be defined as: }
    \begin{equation}
        \mathcal R_{\mathrm{multi}} 
        = D_{\text{JS}}\Bigl(\mathbb P_{\mathcal S}(\boldsymbol\Phi)\,\|\,\mathbb P_{\mathcal T}(\boldsymbol\Phi)\Bigr),
    \end{equation}
    \textit{By Lemma~\ref{appendix:decomposition} (see Appendix), $\mathcal R_{\mathrm{multi}}$ can be upper bounded as: }
    \begin{equation}
    \begin{aligned}
        \mathcal R_{\mathrm{multi}} 
        &\leq 
        \underbrace{D_{\text{JS}}\Bigl(\prod_{m\in \mathcal M} \mathbb P_{\mathcal T}(\phi_m)\,\|\,\mathbb P_{\mathcal T}(\boldsymbol\Phi)\Bigr)}_{\color{gray}\text{Irreducible Risk on Target Domain}} \\
        &+ 
        \underbrace{D_{\text{JS}}\Bigl(\prod_{m\in \mathcal M} \mathbb P_{\mathcal T}(\phi_m)\,\|\,\prod_{m\in \mathcal M} \mathbb P_{\mathcal S}(\phi_m)\Bigr)}_{\color{violet}\text{\textit{Risk 1}: Modality Representation Invariant Risk}} \\
        &+ 
        \underbrace{D_{\text{JS}}\Bigl(\mathbb P_{\mathcal S}(\boldsymbol\Phi)\,\|\,\prod_{m\in \mathcal M} \mathbb P_{\mathcal S}(\phi_m)\Bigr)}_{\color{teal}\text{\textit{Risk 2}: Modality Synergy Invariant Risk}}.
    \end{aligned}
    \label{eq:mm-decomp}
    \end{equation}
    
    Therefore, effective multimodal generalization requires simultaneously minimizing \textit{Risk~1} and \textit{Risk~2}.
\end{definition}

The first term in Eq.~\ref{eq:mm-decomp} is the discrepancy between the product of marginals and the true joint on the \emph{target} domain. 
It quantifies the intrinsic strength of inter-modal dependency on $\mathcal T$, which is irreducible under the source-only training protocol (treated as a constant).

\vspace{1mm}
\noindent\textbf{Modal Representation Invariant Risk \textit{(Risk~1)}}.
The second term compares the products of modality-wise marginals between $\mathcal S$ and $\mathcal T$, reflecting how each modality's representation shifts across domains independently of synergy. 
Under mild conditions, it holds that:
\begin{equation}
\label{eq:r1-sum}
\begin{aligned}
    & D_{\text{JS}}\Bigl(
        \prod_{m\in\mathcal M} \mathbb P_{\mathcal T}(\phi_m)
        \,\Big\|\,
        \prod_{m\in\mathcal M} \mathbb P_{\mathcal S}(\phi_m)
    \Bigr) \\
    &\quad \le \sum_{m\in\mathcal M} 
        D_{\text{JS}}\bigl(
            \mathbb P_{\mathcal T}(\phi_m)
            \,\|\, 
            \mathbb P_{\mathcal S}(\phi_m)
        \bigr) = \sum_{m\in\mathcal M} \mathcal R_{\mathrm{uni}}^{m}.
\end{aligned}
\end{equation}

As discussed in Sec.~\ref{sec:introduction}, this risk arises when models overfit to domain-specific shortcuts within each modality rather than intrinsic spoof cues. 
Mathematically, Eq.~\eqref{eq:r1-sum} demonstrates that \textit{Risk~1} is upper-bounded by the sum of unimodal risks. 
This implies that adding more modalities, while providing richer information, also increases the number of potential domain-specific shortcuts. 
Consequently, the overall generalization risk is amplified in multimodal FAS, which partly explains why multi-modality can sometimes underperform single-modality~\cite{mmdg,xun2025tpami}.

\vspace{1mm}
\noindent\textbf{Modal Synergy Invariant Risk \textit{(Risk~2)}.}
The third term measures the discrepancy between the true joint and the product of marginals on the source domain, capturing how strongly the model can exploit inter-modal co-occurrence patterns during training. 
Risk~2 calls for learning synergy-invariant encoders: we must disentangle intrinsic modality features from cross-modal coincidental statistics.

\subsubsection{Architectural Overview}
As illustrated in Fig.~\ref{fig:method}, RiSe takes multimodal samples from multiple source domains as input. 
Each modality (RGB, depth, and IR) is processed by a dedicated encoder branch, which embeds the raw image into a modality-specific feature representation. 
These features are then passed through separate non-linear projection heads, producing projected modality features in a common latent space. 
The features from all modalities are concatenated and also transformed by a non-linear projection, after which the AsyIRM performs representation learning and live/spoof prediction. 

Before the non-linear projections, the modality-specific features are also fed into the MMSD module for auxiliary training. 
MMSD applies cross-sample frequency mixing and spatial perturbation to enforce disentanglement of intrinsic modal cues from spurious cross-modal correlations. 
The MMSD losses jointly regularize the backbone encoders, guiding them to learn synergy-invariant features that generalize better to unseen domains.

\subsection{Asymmetric Invariant Risk Minimization} \label{sec:asyirm}
AsyIRM is our primary contribution to address \textit{Risk~1}.
It learns a novel geometric representation that is robust to the unique distributional properties of the FAS problem. 
This section first formalizes the underlying asymmetric distribution assumption, then details the two core geometric learning mechanisms of AsyIRM: (1) domain-invariant radial classification and (2) angular domain separation.

\subsubsection{Asymmetric Distribution in FAS} \label{sec:asy}
A foundational premise of our work is the recognition of the inherent distributional asymmetry in the FAS task, 
a characteristic that has also been observed and exploited in prior studies~\cite{ssdg,mmdg,xun2025tpami}.
The ``live'' class ($y\!=\!0$) represents authentic human faces, which, despite variations in identity, pose, and illumination, occupy a relatively compact and well-defined manifold in a suitable feature space. Their features are governed by the consistent biophysical properties of human skin and facial structure. 
In stark contrast, the ``spoof'' class ($y\!=\!1$) is not a single, coherent category but rather a heterogeneous collection of disparate attack schemes. 
These schemes include print attacks, video replays on various screens, and 3D masks made from different materials, each forming its own distinct data distribution. 
Consequently, the ``spoof'' manifold is intrinsically scattered and diverse.
Intuitively, for each genuine face, there exist ``infinitely many'' ways to create spoofs. 
This ``one-to-many'' relation explains why spoof samples span a much broader space compared to the compact live class. 

The asymmetry can be interpreted as: 

\begin{assumption}[Asymmetric Distribution in FAS]
    \label{assum:unimodal_revised}
    For the FAS task, we assume that a well-designed encoder $\phi(\cdot)$ maps live and spoof samples into the embedding space $\mathbb{R}^m$ such that their features are largely decoupled, exhibiting the following key asymmetric characteristics:
    \begin{enumerate}[leftmargin=*]
        \item \textbf{Live samples.} 
        An effective encoder $\phi(\cdot)$, designed for the FAS task, maps real face samples (containing inherently consistent physical and physiological signals) into a compact cluster in the embedding space $\mathbb{R}^d$. 
        We assume this compact cluster is centered at the origin with low variance, and its distribution can be modeled as:
        \begin{equation}
            p(\boldsymbol z\,|\,y=0) = \mathcal{N}(\boldsymbol z\,; \,0, \sigma_0^2 I_d),
        \end{equation}
        where $\sigma_0$ is small.
        
        \item \textbf{Spoof samples.} 
        In contrast, spoof faces form a heterogeneous and scattered distribution, as they can be generated through a wide variety of attack types (e.g., print, replay, or 3D mask). 
        Under mild assumptions (see Appendix for derivation), the spoof class can be modeled as a Gaussian Mixture whose second-moment matrix exhibits a spiked covariance structure: 
        \begin{equation}
            \mathbb{E}\left[\, \boldsymbol z \boldsymbol z^\top \mid y=1 \,\right]
            = \sigma_{\mathrm{eff}}^2 I_d \;+\; \sum_{k=1}^{K}\pi_k\,\mu_k\mu_k^\top,
        \end{equation}
        where $\sigma_{\mathrm{eff}}^2 I$ represents an isotropic variance floor, and the low-rank component $\sum_{k=1}^{K}\pi_k\,\mu_k\mu_k^\top$ corresponds to diverse attack-specific directions, where $\pi_k$ is the prior of attack $k$, and $\mu_k$ is its unique artifact direction. More details are shown in Appendix~\ref{appendix:unimodal_revised}.
    \end{enumerate}
\end{assumption}

The above asymmetric property has a direct implication for conventional hyperplane-based classifiers, such as the commonly used Multilayer Perceptrons (MLPs)~\cite{safas,dadm,mmdg,xun2025tpami}. 
These MLP-based classifiers inherently assume that both classes can be well separated by a linear boundary in the embedding space. 
However, when the live class is compact and isotropic, while the spoof class spans a heterogeneous and scattered distribution, the optimal separating hyperplane becomes highly sensitive to domain-specific variations. 
In particular, spurious shifts in spoof sub-manifolds (e.g., attack types unseen during training) can substantially alter the margin of the hyperplane, thereby enlarging the generalization error bound. 
To better illustrate this issue, we next derive the generalization error bound (based on PAC-Bayes theorem~\cite{pac-bayes}) of IRM under the asymmetric distributional assumption, which motivates AsyIRM.

\begin{lemma}[PAC-Bayes Generalization Error Bound~\cite{pac-bayes}]
With probability at least $1\!-\!\delta$, for a hypothesis $h$ sampled from the posterior distribution $\mathcal{P}$, 
its true risk $\mathcal R_{\mathcal T}(h)$ can be bounded by the empirical risk $\hat{\mathcal R}_{\mathcal S}$ and a KL-divergence term as follows:
\begin{equation}
    \mathcal R_{\mathcal T}(h) \le \hat{\mathcal R}_{\mathcal S}(h) + \sqrt{\frac{\mathrm{KL}(\mathcal{P} \,\|\, \Pi) + \ln(2\sqrt{N}/\delta)}{2N}} ,
\end{equation}
where $N$ is the number of training samples, $\Pi$ is the prior distribution, and $\mathrm{KL}(\mathcal{P} \,\|\, \Pi)$ is the Kullback–Leibler (KL) divergence~\cite{kl} that serves as a key component for this upper bound. 
\label{lemma:bayes}
\end{lemma}

\begin{proposition}[KL Divergence Term of Vanilla IRM]
\label{proposition:sym_kl}
Under Assumption~\ref{assum:unimodal_revised}, for the symmetric IRM framework with a linear classifier, the KL divergence between its posterior distribution $\mathcal P_{\mathrm{sym}}$ and the standard Gaussian prior $\Pi_{\mathrm{sym}}\!=\!\mathcal{N}(\beta; 0, \sigma_\beta^2 I_d)$ is dominated by feature dimensions $d$ and the number of attack types $K$:
\begin{equation}
    \mathrm{KL}(\mathcal{P}_{\mathrm{sym}}\,||\,\Pi_{\mathrm{sym}}) \approx O(d) + O(K).
\end{equation}
Here, $O(d)$ implies that higher-dimensional embeddings introduce more classifier parameters. 
In the PAC-Bayes framework, this directly enlarges the KL term and thus increases the risk of overfitting. 
Meanwhile, $K$ corresponds to the number of spoof attack types: as $K$ grows, the spoof distribution becomes increasingly heterogeneous, further strengthening the asymmetric property of FAS. 
Consequently, both larger $d$ and larger $K$ lead to a looser generalization error bound. 
The proof is shown in Appendix \ref{appendix:sym_kl}.
\end{proposition}

This asymmetry is not merely present but is significantly exacerbated in the multimodal setting. 
Different sensing modalities exhibit varying sensitivities to domain shifts and attack types. 
For example, a change in ambient lighting (a domain shift) will drastically alter the feature representation of a printed photo in the RGB modality but may have a minimal effect on its representation in the depth modality, which primarily captures its planarity. 
Conversely, a change in 3D mask material might be subtle in RGB but create significant new artifacts in the IR spectrum. 
Thus, we extend the above Assumption:

\begin{assumption}[Multimodal Amplification of Asymmetry]
\label{assumption:multimodal_heterogeneous_v2}
Beyond Assumption~\ref{assum:unimodal_revised}, we consider multimodal inputs where the embedding $\boldsymbol z \in \mathbb{R}^{M\times d}$ is a concatenation of $M$ modality-specific embeddings, $\boldsymbol z = [\boldsymbol z_1; \dots; \boldsymbol z_M]$, each with dimension $d$. 
While the spoof class still exhibits a spiked covariance structure, the introduction of multiple modalities amplifies its heterogeneity: 
different modalities incur distinct noise levels due to sensor characteristics and acquisition conditions. 
We therefore model the effective noise floor as a block-diagonal covariance:
\begin{equation}
\small
\sigma_{\mathrm{eff,multi}} = \mathrm{diag}\!\left( \sigma_{\mathrm{eff},1}^2\cdot I_{d},  \sigma_{\mathrm{eff},2}^2\cdot I_{d}, \dots,  \sigma_{\mathrm{eff},M}^2\cdot  I_{d} \right),
\end{equation}
where $\sigma_{\mathrm{eff},m}^2$ is the effective variance for modality $m$. 
The second-moment matrix of spoof samples then becomes:
\begin{equation}
\mathbb{E}\!\left[\, \boldsymbol z \boldsymbol z^\top \mid y=1 \,\right]
= \sigma_{\mathrm{eff,multi}} \;+\; \sum_{k=1}^{K}\pi_k\,\mu_k\mu_k^\top.
\end{equation}
The multimodal spoof distribution is supported over a higher-dimensional space with modality-specific variances, thereby further enlarging the distributional gap between live and spoof classes. A detailed derivation is shown in Appendix~\ref{appendix:multimodal_heterogeneous}.
\end{assumption}

\begin{proposition}[KL Divergence Term in Multimodal IRM]
\label{proposition:multi_kl}
Extending Proposition~\ref{proposition:sym_kl} to the multimodal case, let the multimodal embedding be 
$\boldsymbol z = [\boldsymbol z_1;\dots;\boldsymbol z_M] \in \mathbb{R}^{M \times d}$, 
where $M$ is the number of modalities and $d$ is the dimensionality of each modality-specific feature. 
Under Assumption~\ref{assumption:multimodal_heterogeneous_v2}, 
the KL divergence of the symmetric IRM~\cite{safas,dadm} classifier scales as:
\begin{equation}
    \mathrm{KL}(\mathcal{P}_{\mathrm{sym}}^{\mathrm{multi}}\,||\,\Pi_{\mathrm{sym}}^{\mathrm{multi}}) 
    \;\approx\; O(d\cdot M) \;+\; O(K \cdot \log M).
\end{equation}
The detailed proof is presented in Appendix~\ref{appendix:multi_kl}.
\end{proposition}

In summary, the above analysis shows that multimodal FAS inherently suffers from an enlarged KL term, scaling with both the feature dimension ($O(d\cdot M)$) and the number of attack types ($O(K \cdot \log M)$), which loosens the generalization bound of hyperplane-based IRM. 
This motivates the need for a new formulation that explicitly accommodates the asymmetric distributional property of FAS, leading to our proposed AsyIRM.

\subsubsection{Learning an Invariant Asymmetric Geometry}
To overcome the issues shown in Sec.~\ref{sec:asy}, AsyIRM abandons the notion of a symmetric separator and instead learns a geometric structure tailored to the FAS's topology. 
We decouple the learning of class separation (live vs. spoof) from domain separation by assigning these tasks to different geometric properties of the feature space: the radial magnitude and the angular direction, respectively.

\vspace{1mm}
\noindent\textbf{Domain-Invariant Radial Classifier.} 
Recall from Eq.~\eqref{eq:r1-sum} that the multimodal generalization risk can be upper-bounded by the sum of unimodal risks. 
This motivates us to design per-modality AsyIRM branches, such that each modality $m \in \mathcal M$ is explicitly regularized to minimize its corresponding unimodal risk $\mathcal R_{\mathrm{uni}}^m$. 
Formally, each modality-specific encoder $\phi_m(\cdot)$ generates an embedding, which is then processed by a lightweight projector (Linear–GELU–Linear) to introduce nonlinearity, yielding the final representation $\boldsymbol z_m$.
The projected features are then classified by a modality-specific radial classifier $\omega_m(\cdot)$, which enforces a spherical decision boundary centered at the origin:
\begin{equation}
    \omega_m(\boldsymbol z_m) 
    = 
    \Bigl(\, 
       \underbrace{\bigl(r_m - \|\boldsymbol z_m\|_2\bigr)}_{\text{live logit}}, \quad
       \underbrace{\bigl(\|\boldsymbol z_m\|_2 - r_m\bigr)}_{\text{spoof logit}}
    \,\Bigr),
\end{equation}
where $\omega_m(\cdot)$ outputs a 2-dimensional logit vector for live vs. spoof classification; $r_m$ is the positive radius for classification, also the only trainable parameter within $\omega_m(\cdot)$. 
The classification logits are defined based on the ratio of the feature's norm to this radius, creating a push-pull dynamic relative to the spherical boundary. 
The logit for the ``live'' class is designed to be high when the feature is inside the sphere, while the ``spoof'' logit is high when it is outside.

To explicitly encourage the learned radial classifier to be domain-invariant, we adopt a penalty proposed by IRMv2~\cite{irmv2}. 
Specifically, we encourage the loss landscape with respect to the features to be similar across all domains. 
If the gradients of the loss are aligned across domains, an optimization step that reduces the loss for one domain is likely to do so for all others, preventing the model from learning spurious, domain-specific correlations.
For each source domain $e\in\mathcal E_{\mathcal S}$, we first compute the average gradient $g_m^e$ of the domain-specific radial classification loss $\mathcal L_{\mathrm ce}$ (here we use the cross-entropy loss), with respect to the feature representation $\boldsymbol z^e$ from that domain:
\begin{equation}
    \boldsymbol g^e_m = \mathbb{E}_{(\boldsymbol x_m, y)\sim\mathcal D^e}\Bigl[\nabla_{\boldsymbol z_m} \mathcal{L}_{\mathrm{ce}}\Bigl(\omega_m(\boldsymbol z_m), y\Bigr)\Bigr], 
\end{equation}
and define the modality-wise penalty as the sum of squared Euclidean distances between these average gradient vectors over all pairs of source domains:
\begin{equation}
    \mathcal L^m_{\text{IRM}} = \sum_{e_i, e_j \in \mathcal E_{\mathcal S}, i<j} \|\boldsymbol g^{e_i}_m - \boldsymbol g^{e_j}_m\|_2^2,
\end{equation}
where $\boldsymbol z_m = \phi_m(\boldsymbol x_m)$ is the modality-specific feature. 
Minimizing $\mathcal L^m_{\text{IRM}}$ directly pushes the gradients to align, promoting the learning of a feature representation $\phi_m$ and a radial classifier $r_m$ that constitute a shared optimal solution across all environments.

In addition to per-modality AsyIRM, we introduce a global AsyIRM branch for the final decision making. 
Here, the embeddings from all modalities are concatenated, passed through another projector, and fed into a global radial classifier $\omega_c(\cdot)$. 
This design is motivated by our theoretical analysis (see Sec.~\ref{assumption:multimodal_heterogeneous_v2}), which shows that multimodal inputs further amplify the asymmetric distribution and enlarge the generalization error bound. 
The global classifier thus provides a complementary constraint to reduce the multimodal generalization error risk as a whole. 
Its gradient alignment penalty is defined analogously:
\begin{equation}
    \begin{aligned}
    \boldsymbol g^e_c &= \mathbb{E}_{(\boldsymbol x, y)\sim\mathcal D^e}\Bigl[\nabla_{\boldsymbol z} \mathcal{L}_{\mathrm{ce}}\bigl(\omega_c(\boldsymbol z \bigr), y)\Bigr],\\
    \mathcal L_{\mathrm{IRM}}^c &= \sum_{e_i, e_j \in \mathcal E_{\mathcal S},\, i<j} \|\boldsymbol g_c^{e_i} - \boldsymbol g_c^{e_j}\|_2^2.
    \end{aligned}
\end{equation}

Finally, the overall AsyIRM regularization combines the per-modality and global penalties:
\begin{equation}
    \mathcal L_{\text{IRM}} = \mathcal L^c_{\text{IRM}} + \sum_{m \in \mathcal M}\mathcal L^m_{\text{IRM}}.
\end{equation}

\vspace{2mm}
\noindent\textbf{Angular Separation for Domain Discrimination.} 
Inspired by previous IRM-based works~\cite{safas, dadm}, considering that directly learning domain-invariant representations from scale-limited FAS datasets is inherently difficult, we adopt a similar complementary strategy: rather than suppressing domain information, we preserve it explicitly in the angular subspace, which is orthogonal to the radial decision dimension. 
This design can ensure that domain-specific cues are retained without interfering with the invariant spherical decision boundary.

Concretely, for each modality branch, we introduce a modality-specific angular loss $\mathcal L_{\mathrm{ang}}^m$, and we also employ a global angular loss $\mathcal L_{\mathrm{ang}}^c$ after concatenating all modalities. 
For a batch of L2-normalized features $\{\hat{\boldsymbol z}_i\}_{i=1}^B$ with domain labels $\{e_i\}_{i=1}^B$, we treat samples from the same domain ($e_i\!=\!e_j$) as positive pairs and samples from different domains ($e_i\!\neq\!e_j$) as negative pairs. 
The cosine similarity $\cos(\hat{\boldsymbol z}_i, \hat{\boldsymbol z}_j)$ measures the angular proximity. 
Let $q_{\mathrm{pos}}$ and $q_{\mathrm{neg}}$ be the margins, the angular objective enforces intra-domain compactness and inter-domain separation:
\begin{equation}
\small
\begin{aligned}
    \mathcal L_{\mathrm{ang}}^m
    &= \frac{1}{|Z_{\mathrm{pos}}|}
       \sum_{(i, j) \in Z_{\mathrm{pos}}} 
       \frac{\max\bigl(0,\, q_{\mathrm{pos}} - \cos(\hat{\boldsymbol z}_i, \hat{\boldsymbol z}_j)\bigr)}{\tau} \\
    & + \frac{1}{|Z_{\mathrm{neg}}|}
       \sum_{(i, j) \in Z_{\mathrm{neg}}} 
       \frac{\max\bigl(0,\, \cos(\hat{\boldsymbol z}_i, \hat{\boldsymbol z}_j) - q_{\mathrm{neg}}\bigr)}{\tau},
\end{aligned}
\end{equation}
where $Z_{\mathrm{pos}}$ and $Z_{\mathrm{neg}}$ denote intra-batch positive and negative domain pairs, respectively. 
The final angular loss is the sum of global and modality-specific terms:
\begin{equation}
    \mathcal L_{\mathrm{ang}} = \mathcal L_{\mathrm{ang}}^c + \sum_{m \in \mathcal M} \mathcal L_{\mathrm{ang}}^m.
\end{equation}

Although the angular space explicitly encodes domain identity, this design still induces an invariant representation. 
Since the radial dimension is solely responsible for liveness discrimination, the decision boundary is unaffected by domain-specific variability. 
Meanwhile, the angular separation merely organizes domains into distinct cones, ensuring that embeddings from a new, unseen domain will also be projected into a consistent cone-like structure: live samples remain compact within the decision radius, and spoof samples spread outside the boundary along new angular directions. 
Thus, the classifier in the radial space continues to generalize across domains, while the angular component regularizes representation learning by preventing spurious overlaps between domains.

\vspace{2mm}
\noindent\textbf{Theoretical Generalization Analysis of AsyIRM.} 
We now provide a theoretical analysis of why AsyIRM achieves better generalization than symmetric IRM under the PAC-Bayes framework. 
As shown in Sec.~\ref{proposition:sym_kl}, the KL divergence term of symmetric IRM grows with both the feature dimension $d$ and the number of spoof attack types $K$, leading to an enlarged generalization error bound. 
This result highlights the inherent limitation of hyperplane-based classifiers in the asymmetric setting of FAS, where the spoof class expands rapidly with attack diversity. 

\begin{proposition}[KL Divergence Term of AsyIRM]
\label{proposition:asym_kl}
Under Assumption~\ref{assum:unimodal_revised}, the KL divergence between the posterior distribution $\mathcal{P}_{\mathrm{asym}}$ of the parameters and the prior $\Pi_{\mathrm{asym}}\!\approx\!\mathcal{N}(r; \mu_r, \sigma_r^2)$ is independent of the number of attack types $K$:
\begin{equation}
    \mathrm{KL}(\mathcal{P}_{\mathrm{asym}}\,||\,\Pi_{\mathrm{asym}}) \approx O(1). 
\end{equation}
The detailed proof can be found in Appendix~\ref{appendix:asym_kl}
\end{proposition}

The implication of Proposition~\ref{proposition:asym_kl} is that AsyIRM scales gracefully to high-dimensional multimodal embeddings and diverse spoof categories, without incurring additional KL complexity. 
This directly leads to a tighter PAC-Bayes generalization bound compared to symmetric IRM, as summarized below:

\begin{theorem}[AsyIRM Achieves a Tighter Generalization Error Upper Bound]
\label{theorem:asym_bound}
Consider the PAC-Bayes generalization bound in Lemma~\ref{lemma:bayes}.  
For symmetric IRM with a linear classifier, the KL divergence scales as $\mathrm{KL}(\mathcal{P}_{\mathrm{sym}}\,\|\,\Pi_{\mathrm{sym}}) \approx O(d)+O(K)$, which causes the bound to loosen as either the feature dimension $d$ or the number of spoof attack types $K$ increases.  
In contrast, Proposition~\ref{proposition:asym_kl} shows that AsyIRM maintains $\mathrm{KL}(\mathcal{P}_{\mathrm{asym}}\,\|\,\Pi_{\mathrm{asym}}) \approx O(1)$, independent of $d$ and $K$.  
Consequently, while both bounds follow the same PAC-Bayes form, the asymmetrical formulation eliminates the $d$ and $K$ dependency, leading to a strictly tighter scaling behavior compared to the symmetric case.  
Formally, we have:
\begin{equation}
    \begin{aligned}
        \mathcal R_{\mathrm{sym}}(h) &\;\le\; \hat{\mathcal R}_{\mathrm{sym}}(h) + \tilde{O}\!\left(\sqrt{\tfrac{d+K}{N}}\right),\\
        \mathcal R_{\mathrm{asym}}(h) &\;\le\; \hat{\mathcal R}_{\mathrm{asym}}(h) + \tilde{O}\!\left(\sqrt{\tfrac{1}{N}}\right).
    \end{aligned}
\end{equation}
\end{theorem}

\subsection{Multimodal Synergy Disentanglement (MMSD)}
\label{sec:syndis}

As analyzed in Sec.~\ref{sec:overall}, the multimodal generalization risk decomposes into two terms. 
The previous section addressed the modal representation invariant risk \textit{(Risk~1)} via AsyIRM. 
We now focus on the modal synergy invariant risk \textit{(Risk~2)}, which quantifies the gap between the joint multimodal feature distribution and the product of its unimodal marginals. 
To ensure generalization, both risks must be optimized; this section introduces MMSD, a lightweight self-supervised auxiliary framework that drives feature disentanglement across modalities so as to reduce \textit{Risk~2}.

\subsubsection{Motivation of MMSD}
A central difficulty of \textit{Risk~2} is that multimodal fusion can overfit to domain-specific co-occurrences that do not transfer. 
For example, in RGB-Depth FAS, the shortcut ``Depth flatness + RGB reflection \(\Rightarrow\) spoof'' may hold in training but fails for unseen 3D masks. 
We thus want a representation in which each modality contributes intrinsic spoof cues while spurious cross-modal correlations are discouraged. 
Compared to masked image modeling (e.g., AMA~\cite{ama}), which tends to reconstruct redundant semantics unrelated to spoofness, MMSD adopts targeted, low-overhead pretext tasks that explicitly disrupt spurious synergy both in the frequency domain and in the spatial domain. 
All MMSD operations are applied to modality features $\{\boldsymbol z_m\}$ before the non-linear projector at the end of $\phi_m$ (cf. Fig.~\ref{fig:main}).

In the following, we first introduce cross-sample mixing and perturbation (frequency and spatial), then describe the lightweight decouplers for disentanglement, and finally present a theoretical analysis on how MMSD reduces \textit{Risk~2}.

\subsubsection{Cross-Sample Mixing and Perturbation}
Natural-image frequency components carry complementary information: low frequencies often encode style/illumination/sensor bias and coarse shape, while high frequencies capture edges, texture, and fine detail. 
In FAS, spoof traces may appear in both bands (e.g., print attacks often manifest low-frequency discrepancies due to re-capture, while moiré/edge artifacts are high-frequency). 
Therefore, MMSD does not assume a hard assignment of ``low=style/high=spoof''; instead, it deliberately breaks their co-occurrence patterns by recombining bands across samples and modalities.

\vspace{2mm}
\noindent\textbf{Frequency Decomposition.}
Given a 2D token map $\boldsymbol z \in \mathbb R^{H_t\times W_t}$, its FFT is
\begin{equation}
    \mathcal F(\boldsymbol z)(u,v)=\!\!\sum_{h_t=0}^{H_t-1}\sum_{w_t=0}^{W_t-1}\!\boldsymbol z(h,w)\,e^{-i2\pi(\tfrac{uh_t}{H_t}+\tfrac{vw_t}{W_t})},
\end{equation}
where $\mathcal F(\boldsymbol z)(u,v)$ encodes the spectral response at coordinates $(u,v)$.
The inverse Fast Fourier transform (iFFT), $\mathcal F^{-1}(\cdot)$, reconstructs the spatial domain image.
Following \cite{mfm}, we define a binary mask $\mathcal V\in\{0,1\}^{H_t\times W_t}$ with center $(c_h,c_w)$ and radius $r_f$:
\begin{equation}
    \mathcal V(u,v)=\mathbb I\!\left[d\big((u,v),(c_h,c_w)\big)<r_f\right],
\end{equation}
where $(c_h,c_w)$ denotes the spectrum center; $d(\cdot,\cdot)$ is the Euclidean distance; $r_f$ controls the separation radius.
This masking operation yields a low-pass $\boldsymbol z_l$ and a high-pass component $\boldsymbol z_h$:
\begin{equation}
    \tilde{\boldsymbol z}_l=\mathcal F(\boldsymbol z)\odot\mathcal V,\quad
    \tilde{\boldsymbol z}_h=\mathcal F(\boldsymbol z)\odot(1-\mathcal V),
\end{equation}
with iFFT reconstructions 
$\boldsymbol z_l=\mathcal F^{-1}(\tilde{\boldsymbol z}_l),\;\boldsymbol z_h=\mathcal F^{-1}(\tilde{\boldsymbol z}_h)$.

\vspace{2mm}
\noindent\textbf{Cross-Sample \& Cross-Modality Mixing.}
Let \(\{\boldsymbol z_m^i\}_{i=1}^B\) and \(\{\boldsymbol z_{m'}^j\}_{j=1}^B\) be shuffled mini-batches from two modalities \(m\) and \(m'\), respectively. 
We form mixed features by swapping frequency bands across \emph{different} samples and \emph{different} modalities:
\begin{equation}
\begin{aligned}
    \boldsymbol z_{m\leftarrow m'}^{i\leftarrow j}
    &= \mathcal F^{-1}\!\big(\tilde{\boldsymbol z}_{m,l}^i + \tilde{\boldsymbol z}_{m',h}^j\big),\\
    \boldsymbol z_{m'\leftarrow m}^{j\leftarrow i}
    &= \mathcal F^{-1}\!\big(\tilde{\boldsymbol z}_{m',l}^j + \tilde{\boldsymbol z}_{m,h}^i\big).
\end{aligned}
\end{equation}

By drawing low/high components from independent sources (different samples and often different domains/identities/labels), MMSD destroys the spurious co-occurrence statistics that a model could otherwise memorize. 
Intuitively, under such mixing, the only stable signal is the intrinsic band-specific cue of each modality, which the encoder must capture to succeed in downstream self-supervision.

\subsubsection{Random Sampling and Spatial Perturbation}
We further apply token-level random sampling and spatial shuffling to prevent positional shortcuts and residual alignment bias. 
At each location \(p\), a token is sampled from four candidates:
\begin{equation}
    \hat{\boldsymbol z}^p_{m, m'} \sim \mathcal U\!\big\{
    \boldsymbol z_m^i,\ \boldsymbol z_{m'}^j,\
    \boldsymbol z_{m\leftarrow m'}^{i\leftarrow j},\
    \boldsymbol z_{m'\leftarrow m}^{j\leftarrow i}
    \big\}.
\end{equation}

Then a random permutation $\pi$ shuffles token order. 
The motivation behind $\pi$ is that, across modalities, face parts share consistent relative geometry, which can inadvertently leak inter-modal alignment cues. 
Spatial shuffling removes these shortcuts so that recognizing token origin (and later recovering position) requires learning local, spoof-relevant structure rather than global layout.

\subsubsection{Lightweight Decoupler for Disentanglement}
The mixed and perturbed tokens are fed into lightweight modality-specific decouplers, 
formally denoted as $\{f_{\mathrm{dec}}^{m}\}_{m\in\mathcal M}$. 
Each $f_{\mathrm{dec}}^{m}$ consists of two auxiliary heads, a low-frequency head $f_{l}^{m}$ and a high-frequency head $f_{h}^{m}$, both implemented as small MLP classifiers. 
These paired heads provide fine-grained supervision for disentangling the frequency components within each modality.

\vspace{2mm}
\noindent\textbf{Frequency-origin identification.}
Given a mixed token $\hat{\boldsymbol z}_p$, the task of these heads is to identify the source modality of its \emph{low}- and \emph{high}-frequency components.  
Formally, let $y_{l}(p)/y_{h}(p) \in \mathcal M$ denote the ground-truth source modalities of the low/high parts of position $p$ (determined by the mixing recipe).  
Since mixing is performed across all modality pairs, the decoupler $f_{\mathrm{dec}}^{m}$ disentangles tokens originating from its own modality $m$ and from every other modality $m'\in\mathcal M,\, m'\neq m$. Let $P_t = H_t \times W_t$. The per-modality identification losses are:
\begin{equation}
\small
\begin{aligned}
    \mathcal L_{l}^m
    &= \sum_{m'\in\mathcal M, m'\neq m}\Big[-\frac{1}{P_t}\sum_{p=1}^{P_t}\mathcal L_{\mathrm{ce}}\big(f_l^m(\hat{\boldsymbol z}^p_{m,m'}), y_{l}(p)\big)\Big], \\
    \mathcal L_{h}^m
    &= \sum_{m'\in\mathcal M, m'\neq m}\Big[-\frac{1}{P_t}\sum_{p=1}^{P_t}\mathcal L_{\mathrm{ce}}\big(f_h^m(\hat{\boldsymbol z}^p_{m,m'}), y_{h}(p)\big)\Big].
\end{aligned}
\end{equation}

Therefore, the total frequency-origin loss for modality $m$ aggregates over all other modalities:  
\begin{equation}
    \mathcal L_{\mathrm{freq}}^m \;=\; \mathcal L_{l}^m + \mathcal L_{h}^m.
\end{equation}

Optimizing this loss ensures that each decoupler learns to recover intrinsic frequency-specific cues in a pairwise manner across all modality combinations, effectively suppressing spurious cross-modal co-occurrence patterns.

\vspace{2mm}
To further disrupt spurious correlations, we randomly permute the spatial order of tokens and require the model to recover their original positions. 
For each token $p$, let $y_{\mathrm{pos}}^p$ denote its ground-truth position label prior to permutation. 
Given the shuffled token $\pi(\hat{\boldsymbol z}^p)$, a lightweight position head $f_{\mathrm{pos}}$ regresses its original position. 
The loss is defined as:
\begin{equation}
\small
    \mathcal L_{\mathrm{pos}}
    = \sum_{m,m'\in\mathcal M, m'\neq m}\Big[\frac{1}{P_t}\sum_{p=1}^{P_t}
      \big\| f_{\mathrm{pos}}\big(\pi(\hat{\boldsymbol z}_{m, m'}^p)\big) - y_{\mathrm{pos}}^p \big\|_2^2\Big].
\end{equation}

This regression task forces the encoder to retain spatially consistent facial structure cues that generalize across domains, rather than memorizing domain-specific context.

\vspace{2mm}
\noindent\textbf{Total Auxiliary Objective.}
The final MMSD loss aggregates modality-wise frequency supervision and the position loss:
\begin{equation}
\mathcal L_{\text{MMSD}}
= \mathcal L_{\text{pos}} + \sum_{m\in\mathcal M}\mathcal L_{\text{freq}}^m.
\end{equation}

\subsubsection{Theoretical Analysis on MMSD}
\begin{recall}
\textit{The modal synergy invariant risk (Risk~2) measures the departure of the joint feature distribution from the product of unimodal marginals (cf. Eq.~\ref{eq:mm-decomp}):}
\begin{equation}
\mathcal R_{\mathrm{syn}}
= D_{\text{JS}}\Big(\mathbb P_{\mathcal S}(\boldsymbol\Phi)\,\Big\|\,\prod_{m\in\mathcal M}\mathbb P_{\mathcal S}(\phi_m)\Big).
\end{equation}
\end{recall}

Our mixing draws low/high components from independent sources across samples/modalities, so the target distribution for an ideal disentangled encoder is close to the factorized one. 
The following proposition links MMSD to a decrease in \textit{Risk~2}.

\begin{proposition}[MMSD Reduces Modal Synergy Risk]
\label{prop:mmsd_main_body}
Let the decouplers be sufficiently expressive to solve the auxiliary tasks. If the encoder $\Phi$ is optimized to minimize the MMSD loss $\mathcal L_{\text{MMSD}}$ over the distribution of cross-sample mixed features, it is incentivized to learn a representation where features from each modality are self-contained and identifiable without relying on statistical co-occurrence with others. Consequently, the learned joint feature distribution is driven to approximate a factorized distribution:
\begin{equation}
\mathbb P_{\mathcal S}(\boldsymbol\Phi) \;\to\; \prod_{m\in\mathcal M}\mathbb P_{\mathcal S}(\phi_m),
\end{equation}
which in turn minimizes the modal synergy invariant risk, $\mathcal R_{\mathrm{syn}}$. A detailed proof is provided in Appendix~\ref{appendix:mmsd_main_body}.
\end{proposition}

\begin{remark}[Multimodal Synergy is Not Free]
    While multimodal fusion offers richer cues, Proposition~\ref{proposition:multi_kl} shows that it also amplifies the generalization error bound \textit{(Risk~1)} by enlarging distributional asymmetries. 
    Similarly, cross-domain multimodal learning introduces the modal synergy invariant risk \textit{(Risk~2)}, as models may exploit domain-specific inter-modal co-occurrences that fail to generalize. 
\end{remark}

\begin{table*}[t]
\centering
\caption{Overview of benchmark protocols for multimodal cross-domain FAS~\cite{mmdg,xun2025tpami}. Each protocol is designed to simulate a specific deployment challenge.}
\vspace{-0.8em}
\label{tab:protocols}
\renewcommand{\arraystretch}{1.2} 
\setlength{\tabcolsep}{8pt} 
\begin{tabularx}{\linewidth}{>{\raggedright\arraybackslash}p{3.0cm}| 
                                    >{\raggedright\arraybackslash}X 
                                    >{\raggedright\arraybackslash}X}
\toprule
\textbf{Protocol Name} & \textbf{Setup / Details} & \textbf{Motivation \& Focus} \\
\midrule
\textbf{Protocol 1: Complete modalities} & 
Multi-dataset leave-one-out (LOO). Train on three datasets and test on the held-out one, e.g., \textbf{CPS}$\rightarrow$\textbf{W}. & 
Simulates cross-domain deployment when all modalities are available, focusing on generalization to unseen domains. \\

\textbf{Protocol 2: Missing modalities (test-time)} & 
Extend Protocol~1 but drop one or more modalities (D, I, or both) at test time. & 
Evaluates robustness against sensor failures or network issues that cause missing modalities during deployment. \\

\textbf{Protocol 2+: Missing modalities (train-time)} & 
During training, each sample's D/IR modalities are randomly dropped with 70\% probability. & 
Simulates incomplete multimodal datasets, testing the model's ability to learn robustly under missing training modalities. \\

\textbf{Protocol 3: Limited source domains} & 
Restrict training to fewer source domains (datasets), e.g., \textbf{CW}$\rightarrow$\textbf{PS} or \textbf{PS}$\rightarrow$\textbf{CW}. & 
Evaluates generalization under data-scarce conditions, focusing on resource-constrained scenarios. \\
\bottomrule
\end{tabularx}
\end{table*}

\begin{table*}[t]
    \label{tab:protocol1}
    \caption{Cross-dataset testing results~(\%) under the complete-modal scenarios (\textbf{Protocol 1}) among CASIA-CeFA (\textbf{C}), PADISI (\textbf{P}), CASIA-SURF (\textbf{S}), and WMCA (\textbf{W}). The best and second-best results are in \textbf{bold} and \underline{underline}, respectively}
     \vspace{-0.8em}
	\setlength{\tabcolsep}{6pt}
          \begin{tabularx}{\linewidth}{l *{10}{>{\centering\arraybackslash}X}}
            \toprule[1.2pt]
            \multicolumn{1}{l|}{\multirow{2}{*}{\textbf{Method}}} & \multicolumn{2}{c}{\textbf{CPS $\rightarrow$ W}} & \multicolumn{2}{c}{\textbf{CPW $\rightarrow$ S}} & \multicolumn{2}{c}{\textbf{CSW $\rightarrow$ P}} & \multicolumn{2}{c}{\textbf{PSW $\rightarrow$ C}} & \multicolumn{2}{c}{\textbf{Average}} \\ 
            \cmidrule(lr){2-3}\cmidrule(lr){4-5}\cmidrule(lr){6-7}\cmidrule(lr){8-9}\cmidrule(lr){10-11}
            \multicolumn{1}{l|}{} & HTER $\downarrow$ & AUC $\uparrow$ & HTER $\downarrow$ & AUC $\uparrow$ & HTER $\downarrow$ & AUC $\uparrow$ & HTER $\downarrow$ & AUC $\uparrow$ & HTER $\downarrow$ & AUC $\uparrow$ \\ 
            \midrule[1.2pt]
            \multicolumn{1}{l|}{SSDG~\cite{ssdg}} & 26.09 & 82.03 & 28.50 & 75.91 & 41.82 & 60.56 & 40.48 & 62.31 & 34.22 & 70.20 \\
            \multicolumn{1}{l|}{SSAN~\cite{ssan}} & 17.73 & 91.69 & 27.94 & 79.04 & 34.49 & 68.85 & 36.43 & 69.29 & 29.15 & 77.22 \\
            \multicolumn{1}{l|}{IADG~\cite{iadg}} & 27.02 & 86.50 & 23.04 & 83.11 & 32.06 & 73.83 & 39.24 & 63.68 & 30.34 & 76.78 \\
            \multicolumn{1}{l|}{SA-FAS~\cite{safas}} & 23.04 & 83.11 & 32.06 & 73.83 & 39.24 & 63.68 & 30.34 & 76.78 & 31.17 & 74.28 \\
            \multicolumn{1}{l|}{ViTAF~\cite{adaptive-transformer}} & 20.58 & 85.82 & 29.16 & 77.80 & 30.75 & 73.03 & 39.75 & 63.44 & 30.06 & 75.02 \\
            \multicolumn{1}{l|}{MM-CDCN~\cite{mm-cdcn}} & 38.92 & 65.39 & 42.93 & 59.79 & 41.38 & 61.51 & 48.14 & 53.71 & 42.84 & 60.10 \\
            \multicolumn{1}{l|}{CMFL~\cite{cmfl}} & 18.22 & 88.82 & 31.20 & 75.66 & 26.68 & 80.85 & 36.93 & 66.82 & 28.26 & 78.04 \\
            \multicolumn{1}{l|}{AMA~\cite{ama}} & 17.56 & 88.74 & 27.50 & 80.00 & 21.18 & 85.51 & 47.48 & 55.56 & 28.43 & 77.45 \\
            \multicolumn{1}{l|}{VP-FAS~\cite{vp-fas}} & 16.26 & 91.22 & 24.42 & 81.07 & 21.76 & 85.46 & 39.35 & 66.55 & 25.45 & 81.08 \\
            \multicolumn{1}{l|}{FLIP~\cite{flip}} & 13.19 & 93.79 & 11.73 & 94.93 & 17.39 & 90.63 & 22.14 & 83.95 & 16.11 & 90.83 \\ \midrule
            \multicolumn{1}{l|}{MMDG~\cite{mmdg}} & 12.79 & 93.83 & 15.32 & 92.86 & 18.95 & 88.64 & 29.93 & 76.52 & 19.25 & 87.96 \\ 
            \multicolumn{1}{l|}{DADM~\cite{dadm}} & 11.71 & 94.89 & \underline{6.92} & 97.66 & 19.03 & 88.22 & \textbf{16.87} & \underline{91.08} & 13.63 & 92.96 \\ 
            \multicolumn{1}{l|}{MMDG++~\cite{xun2025tpami}} & \underline{2.08} & \underline{99.82} & 8.72 & \underline{96.77} & \underline{10.24} & \underline{94.97} & 18.87 & 89.28 & \underline{9.98} & \underline{95.21} \\ 
            \rowcolor{lightblue!70}
            \multicolumn{1}{l|}{\textbf{RiSe~(Ours)}} & \textbf{0.89} & \textbf{99.96} & \textbf{5.32} & \textbf{98.56} & \textbf{7.64} & \textbf{96.85} & \underline{16.93} & \textbf{91.37} & \textbf{7.70} & \textbf{96.69} \\ 
            \bottomrule[1.2pt]
        \end{tabularx}
    \label{tab:p1}
\end{table*}

\section{Experiments} \label{sec:experiment}
\subsection{Implementation Details}
All RGB, depth, and infrared inputs are resized to $224\!\times224\!\times3$. Each modality is tokenized into a $14\!\times\!14$ patch sequence with an additional class token, where the default embedding dimension of each token is set to 768. 
The backbone is initialized from a pretrained CLIP~\cite{clip} with the vision encoder, and fine-tuned via parameter-efficient transfer learning (PETL)~\cite{s-adapter,rizhao2025tpami,adaptive-transformer,mmdg,xun2025tpami,dadm} using lightweight convolutional adapters (ConvPass~\cite{convpass}) with hidden dimension 8. 
Missing-modality simulation is achieved by zeroing out the corresponding modality input.
For non-linear projectors, features are first projected to one-quarter of their dimension, activated by GELU, and then mapped back to the original hidden size. 
For frequency decomposition, the cutoff radius $r_f$ is empirically set to 1. 
The initial radius in AsyIRM is parameterized as $\mathrm{softplus}(0.0)$ to ensure positive.
Training is performed with Adam, with a learning rate of $5\!\times\!10^{-5}$, weight decay of $1\!\times\!10^{-3}$, and batch size of 48. Sampling is balanced across both classes and source domains. 
RiSe is trained for 300 epochs. 
To stabilize training, we introduce an auxiliary loss $\mathcal L_{\mathrm{aux}}$, implemented as a linear classifier applied to CLIP-derived features before the nonlinear projector and AsyIRM. 
This auxiliary branch enforces linear separability at an early stage, easing optimization of the subsequent asymmetric manifold. 
The training objective is:
\begin{equation}
    \mathcal L = \mathcal L_{\mathrm{CLS}} + \lambda_{\mathrm{1}}\mathcal L_{\mathrm{IRM}} + \lambda_{\mathrm{2}}\mathcal L_{\mathrm{ang}} + \lambda_{\mathrm{3}}\mathcal L_{\mathrm{MMSD}} + \lambda_{\mathrm{4}}\mathcal L_{\mathrm{aux}},
    \label{eq:loss-weight}
\end{equation}
where the weights of loss terms $\lambda_{1}\sim\lambda_{4}$ are empirically set to 0.5, 0.5, 1.0, and 1.0, respectively. 

\subsection{Multi-Modal Cross-Domain Benchmark} \label{sec:bench}
To make comprehensive evaluations, we follow the large-scale benchmark adopted in~\cite{mmdg,xun2025tpami,dadm}, which involves four prominent multimodal FAS datasets, i.e., CASIA-CeFA (\textbf{C})~\cite{cefa}, PADISI-Face (\textbf{P})~\cite{padisi}, CASIA-SURF (\textbf{S})~\cite{mmfa}, and WMCA (\textbf{W})~\cite{wmca}, and defines four cross-domain evaluation protocols. 
Half Total Error Rate (HTER) and Area Under Curve (AUC) are used as evaluation metrics. 
As shown in Table~\ref{tab:protocols}, this benchmark includes four protocols, each designed to simulate a specific deployment challenge.

\subsection{Cross-Domain Testing}
We compare our approach with three categories of open-source baselines:
\begin{enumerate}
    \item \textbf{Unimodal DG methods:} SSDG~(CVPR'20)~\cite{ssdg}, SSAN~(CVPR'22)~\cite{ssan}, IADG~(CVPR'23)~\cite{iadg}, ViTAF~(ECCV'22)~\cite{adaptive-transformer}, FLIP~(ICCV'23)~\cite{flip}, and SA-FAS~(CVPR'23)~\cite{safas};

    \item \textbf{Multimodal non-DG methods:} CMFL~(CVPR'21)~\cite{cmfl}, MM-CDCN~(CVPRW'20)~\cite{mm-cdcn}, AMA~(IJCV'24)~\cite{ama}, and VP-FAS~(TDSC'24)~\cite{vp-fas};

    \item \textbf{Multimodal DG methods:} MMDG~(CVPR'24)~\cite{mmdg}, DADM~(ICCV'25)~\cite{dadm}, and MMDG++~(TPAMI'25)~\cite{xun2025tpami}.
\end{enumerate}

Notably, DADM~\cite{dadm} and SA-FAS~\cite{safas} are IRM-based methods.
For methods not originally designed for multimodal FAS, we adapt their architectures to a consistent evaluation framework. 
Specifically, we adopt the same multi-branch encoder with late fusion as RiSe whenever feasible; otherwise, we employ early fusion by concatenating modality inputs along the channel dimension. 
Hyperparameters are either taken from the official implementations or tuned for optimal performance under our benchmark. 
All methods are retrained strictly following the proposed benchmark protocols to ensure fairness.

\begin{table*}[t]
    \caption{Cross-dataset testing results~(\%) under the scenarios with missing modalities at inference stage (\textbf{Protocol 2}) among CASIA-CeFA (\textbf{C}), PADISI (\textbf{P}), CASIA-SURF (\textbf{S}), and WMCA (\textbf{W}). The best and second-best results are in \textbf{bold} and \underline{underline}, respectively}
    \vspace{-0.8em}
    \centering
    \begin{tabularx}{\linewidth}{l| *{8}{>{\centering\arraybackslash}X}}
        \toprule[1.2pt]
        \multirow{2}{*}{\textbf{Method}} & \multicolumn{2}{c}{\textbf{Missing DEP}} & \multicolumn{2}{c}{\textbf{Missing IR}} & \multicolumn{2}{c}{\textbf{Missing DEP \& IR}} & \multicolumn{2}{c}{\textbf{Average}} \\ \cmidrule(l){2-9} 
         & HTER $\downarrow$ & AUC $\uparrow$ & HTER $\downarrow$ & AUC $\uparrow$ & HTER $\downarrow$ & AUC $\uparrow$ & HTER $\downarrow$ & AUC $\uparrow$ \\ 
         \midrule[1.2pt]
        SSDG~\cite{ssdg} & 38.92 & 65.45 & 37.64 & 66.57 & 39.18 & 65.22 & 38.58 & 65.75 \\
        SSAN~\cite{ssan} & 36.77 & 69.21 & 41.20 & 61.92 & 33.52 & 73.38 & 37.16 & 68.17 \\
        IADG~\cite{iadg} & 40.72 & 58.72 & 42.17 & 61.83 & 37.50 & 66.90 & 40.13 & 62.49 \\
        SA-FAS~\cite{safas} & 36.30 & 69.07 & 39.80 & 62.69 & 33.08 & 74.29 & 36.40 & 68.68 \\
        ViTAF~\cite{adaptive-transformer} & 34.99 & 73.22 & 35.88 & 69.40 & 35.89 & 69.61 & 35.59 & 70.64 \\
        MM-CDCN~\cite{mm-cdcn} & 44.90 & 55.35 & 43.60 & 58.38 & 44.54 & 55.08 & 44.35 & 56.27 \\
        CMFL~\cite{cmfl} & 31.37 & 74.62 & 30.55 & 75.42 & 31.89 & 74.29 & 31.27 & 74.78 \\
        AMA~\cite{ama} & 29.25 & 77.70 & 32.30 & 74.06 & 31.48 & 75.82 & 31.01 & 75.86 \\
        VP-FAS~\cite{vp-fas} & 29.13 & 78.27 & 29.63 & 77.51 & 30.47 & 76.31 & 29.74 & 77.36 \\
        FLIP~\cite{flip} & 23.66 & 83.90 & 24.06 & 84.04 & 27.07 & 79.79 & 27.93 & 79.44 \\ \midrule
        MMDG~\cite{mmdg} & 24.89 & 82.39 & 23.39 & 83.82 & 25.26 & 81.86 & 24.51 & 82.69 \\ 
        DADM~\cite{dadm} & 21.56 & 85.17 & 20.82 & 85.28 & 22.61 & 84.04 & 21.66 & 84.83 \\ 
        MMDG++~\cite{xun2025tpami} & \underline{15.11} & \underline{92.01} & \underline{15.56} & \underline{91.05} & \underline{17.64} & \underline{89.51} & \underline{16.10} & \underline{90.85} \\ 
        \rowcolor{lightblue!70}
        \textbf{RiSe~(Ours)} & \textbf{13.52} & \textbf{92.17} & \textbf{11.53} & \textbf{94.55} & \textbf{15.52} & \textbf{90.39} & \textbf{13.52} & \textbf{92.37} \\
        \bottomrule[1.2pt]
    \end{tabularx}
    \label{tab:p2}
\end{table*}
\begin{table*}[t]
    \caption{Cross-dataset testing results~(\%) under the scenarios with missing modalities during training (\textbf{Protocol 2+}) among CASIA-CeFA (\textbf{C}), PADISI (\textbf{P}), CASIA-SURF (\textbf{S}), and WMCA (\textbf{W}). The best and second-best results are in \textbf{bold} and \underline{underline}, respectively}
    \vspace{-0.8em}
      \begin{tabularx}{\linewidth}{l| *{10}{>{\centering\arraybackslash}X}}
        \toprule[1.2pt]
        \multirow{2}{*}{\textbf{Method}} & \multicolumn{2}{c}{\textbf{CPS $\rightarrow$ W}} & \multicolumn{2}{c}{\textbf{CPW $\rightarrow$ S}} & \multicolumn{2}{c}{\textbf{CSW $\rightarrow$ P}} & \multicolumn{2}{c}{\textbf{PSW $\rightarrow$ C}} & \multicolumn{2}{c}{\textbf{Average}} \\
        \cmidrule(lr){2-3}\cmidrule(lr){4-5}\cmidrule(lr){6-7}\cmidrule(lr){8-9}\cmidrule(lr){10-11}
         & HTER $\downarrow$ & AUC $\uparrow$ & HTER $\downarrow$ & AUC $\uparrow$ & HTER $\downarrow$ & AUC $\uparrow$ & HTER $\downarrow$ & AUC $\uparrow$ & HTER $\downarrow$ & AUC $\uparrow$ \\ 
        \midrule[1.2pt]
        SSDG~\cite{ssdg} & 28.97 & 77.48 & 31.01 & 75.57 & 45.12 & 59.64 & 44.18 & 60.32 & 37.32 & 68.25 \\
        SSAN~\cite{ssan} & 30.68 & 76.25 & 34.33 & 71.73 & 37.35 & 70.00 & 38.98 & 65.94 & 35.34 & 70.98 \\
        IADG~\cite{iadg} & 36.87 & 67.14 & 24.53 & 81.99 & 49.04 & 52.30 & 48.88 & 50.36 & 39.83 & 62.95 \\
        ViTAF~\cite{adaptive-transformer} & 26.09 & 81.07 & 22.61 & 85.29 & 41.38 & 60.51 & 45.48 & 59.30 & 33.89 & 71.54 \\
        MM-CDCN~\cite{mm-cdcn} & 44.64 & 56.43 & 47.41 & 53.15 & 47.68 & 50.06 & 47.51 & 54.06 & 46.81 & 53.43 \\
        CMFL~\cite{cmfl} & 21.73 & 87.06 & 32.50 & 72.90 & 31.19 & 75.29 & 38.61 & 65.01 & 31.01 & 75.07 \\
        AMA~\cite{ama} & 18.92 & 88.79 & 33.06 & 72.39 & 22.78 & 86.76 & 35.11 & 71.47 & 27.47 & 79.85 \\
        VP-FAS~\cite{vp-fas} & 22.40 & 86.95 & 26.60 & 81.67 & 23.85 & 80.30 & 46.43 & 57.55 & 29.82 & 76.62 \\
        FLIP~\cite{flip} & 17.96 & 90.64 & 15.00 & 92.77 & 25.20 & 83.96 & 24.40 & 83.08 & 20.64 & 87.61 \\ 
        \midrule
        MMDG~\cite{mmdg} & 17.66 & 89.94 & 19.32 & 87.80 & 21.46 & 86.26 & 33.27 & 72.75 & 22.93 & 84.19 \\ 
        DADM~\cite{dadm} & 15.04 & 92.07 & 17.33 & 89.09 & 19.76 & 88.84 & \underline{22.02} & \underline{87.16} & 18.54 & 89.29 \\ 
        MMDG++~\cite{xun2025tpami} & \underline{8.68} & \underline{97.54} & \underline{11.23} & \textbf{96.03} & \underline{15.88} & \underline{93.32} & 24.60 & 79.39 & \underline{15.10} & \underline{91.57} \\ 
        \rowcolor{lightblue!70}
        \textbf{RiSe~(Ours)} & \textbf{6.39} & \textbf{98.85} & \textbf{11.03} & \underline{94.20} & \textbf{10.80} & \textbf{94.36} & \textbf{20.18} & \textbf{89.11} & \textbf{12.10} & \textbf{94.13} \\ 
        \bottomrule[1.2pt]
    \end{tabularx}
    \label{tab:p3}
\end{table*}
\subsubsection{Results on Protocol 1: Complete Modalities}
Under the complete-modal cross-dataset setting, RiSe achieves a decisive lead over all baselines. 
With all modalities available, RiSe reports an average HTER of 7.70\% and AUC of 96.69\%, substantially outperforming unimodal DG methods (e.g., SSDG, SSAN, and SA-FAS with much higher errors) and even CLIP-based FLIP ($\sim$16.1\% HTER). 
Conventional multimodal methods (e.g., AMA, VP-FAS, and CMFL) also underperform, often exceeding 27\% HTER. 
Compared with advanced multimodal DG baselines, RiSe still shows clear gains: DADM and MMDG++ achieve HTER around 13–15\% and AUC $\sim$91–93\%, whereas RiSe nearly halves the error and raises AUC by $\sim$6\%. 
These improvements stem from AsyIRM, which enforces invariant radial decisions and reduces cross-domain KL divergence, and MMSD, which disentangles modality cues to avoid spurious synergy. 
Together, they enable RiSe to consistently capture genuine liveness cues, delivering SoTA results across all LOO sub-protocols.
Under the complete-modal cross-dataset setting, RiSe achieves a decisive performance lead over all baselines. 
With all modalities available, RiSe's average HTER is only 7.70\%, and AUC reaches 96.69\%, dramatically outperforming prior methods. 
In contrast, unimodal DG approaches like SSDG, SSAN, and SA-FAS struggle with much higher error rates, highlighting the value of multi-modal input. 
Even the CLIP-based~\cite{clip} unimodal method (FLIP) manages an average HTER of $\sim$16.1\%, far above RiSe. 
Traditional multi-modal (non-DG) methods show moderate improvements but still falter on unseen domains (e.g., AMA, VP-FAS, and CMFL obtain HTER$\approx$27\% on average, AUC$<$82\%). 
In contrast, RiSe maintains low errors across all LOO tests. 
For instance, on \textbf{PSW$\rightarrow$C}, RiSe attains 16.93\% HTER (nearly halving the $\sim$33–48\% HTER of many baselines). 
Compared to advanced multimodal DG competitors, RiSe still shows clear gains. 
The competitive previous multimodal DG methods (e.g., DADM and MMDG++) report average HTER around 13–15\% and AUC $\sim$91–93\% in Protocol 1, whereas RiSe drives the HTER down to 7.70\% and pushes AUC up to 96.69\%. 

\subsubsection{Results on Protocol 2: Test-Time Modality Missing}
In Protocol 2, we examine RiSe's robustness when depth and/or infrared modalities are missing during inference. 
RiSe consistently outperforms all baselines, maintaining high accuracy without retraining. 
For example, when depth is missing at test (``Missing DEP''), RiSe's HTER is 13.52\% with AUC 92.17\%, compared to 25-40\% HTER obtained by most baselines. Even IRM-based DADM shoots up to $\sim$21.6\% HTER without depth, underscoring how heavily others depended on the depth modality, while prior SoTA (i.e., MMDG++) records 15.11\%/90.85\%. 
When infrared is missing (``Missing IR''), RiSe's performance (11.53\% HTER, 94.55\% AUC) remains close to full-modality results (7.70\%/96.69\%), highlighting graceful degradation. 
Even in the most challenging RGB-only case, RiSe yields 15.52\% HTER, dramatically outperforming FLIP (27.09\%). 
This resilience may stem from MMSD's ability to disentangle and fuse modality-specific features without relying on brittle inter-modal correlations, ensuring each stream contributes robustly even in isolation. 
Within RiSe, MMSD requires RGB, depth, and infrared modalities to be trained to provide standalone informative features (through cross-sample and cross-modality mixing), so at test-time each modality can ``stand on its own'' if needed.

\subsubsection{Results on Protocol 2+: Train-Time Modality Missing}
Protocol~2+ examines a more challenging setting of training with incomplete data, where DEP and IR modalities are randomly dropped with a 70\% probability.
Under these stringent conditions, RiSe demonstrates superior robustness, achieving the best average performance with a 12.10\% HTER and 94.13\% AUC. This result significantly outperforms strong baselines like MMDG++ (15.10\% HTER) and DADM (18.54\% HTER). The advantage is particularly stark in difficult scenarios such as \textbf{PSW$\rightarrow$C}, where RiSe maintains a 20.18\% HTER, while other methods like MMDG degrade severely to 33.27\%.
These results indicate that RiSe's resilience may stem from its core components. 
AsyIRM drives the model to learn modality-invariant features that are effective even when data streams are absent. Simultaneously, MMSD acts as a form of modality augmentation, training the model to not overfit to specific cross-modal correlations.
Consequently, these designs yield a robust representation that is highly tolerant of missing training data.

\subsubsection{Results on Protocol 3: Limited Source Domains}
\begin{table}[t]
  \caption{Cross-dataset testing results~(\%) under the limited source domain scenarios (\textbf{Protocol 3}) among CASIA-CeFA (\textbf{C}), PADISI-USC (\textbf{P}), CASIA-SURF (\textbf{S}), and WMCA (\textbf{W}).}
  \vspace{-0.8em}
  \begin{tabularx}{\linewidth}{l| *{4}{>{\centering\arraybackslash}X}} 
    \toprule[1.2pt]
    \multicolumn{1}{l|}{\multirow{2}{*}{\textbf{Method}}} & \multicolumn{2}{c}{\textbf{CW} $\rightarrow$ \textbf{PS}}
    & \multicolumn{2}{c}{\textbf{PS} $\rightarrow$ \textbf{CW}} \\
    \cmidrule(lr){2-3}\cmidrule(lr){4-5}
    & HTER $\downarrow$ & AUC $\uparrow$ & HTER $\downarrow$ & AUC $\uparrow$ \\
    \midrule[1.2pt]
    SSDG~\cite{ssdg} & 25.34 & 80.17 & 46.98 & 54.29 \\
    SSAN~\cite{ssan} & 26.55 & 80.06 & 39.10 & 67.19 \\
    IADG~\cite{iadg} & 22.82 & 83.85 & 39.70 & 63.46 \\
    SA-FAS~\cite{safas} & 25.20 & 81.06 & 36.59 & 70.03 \\
    ViTAF~\cite{adaptive-transformer} & 29.64 & 77.36 & 39.93 & 61.31 \\
    MM-CDCN~\cite{mm-cdcn} & 29.28 & 76.88 & 47.00 & 51.94 \\
    CMFL~\cite{cmfl} & 31.86 & 72.75 & 39.43 & 63.17 \\
    AMA~\cite{ama} & 29.25 & 76.89 & 38.06 & 67.64 \\
    VP-FAS~\cite{vp-fas} & 25.90 & 81.79 & 44.37 & 60.83 \\
    FLIP~\cite{flip} & 15.92 & 92.38 & 23.85 & 83.46 \\
    \midrule
    MMDG~\cite{mmdg} & 20.12 & 88.24 & 36.60 & 70.35 \\
    DADM~\cite{dadm} & 12.61 & 93.81 & 20.40 & 89.51 \\
    MMDG++~\cite{xun2025tpami} & 10.67 & 95.95 & 21.55 & 86.73 \\
    \rowcolor{lightblue!70}
    \textbf{RiSe~(Ours)} & \textbf{8.05} & \textbf{96.63} & \textbf{19.46} & \textbf{89.55} \\
    \bottomrule[1.2pt]
  \end{tabularx}
  \label{tab:p4}
\end{table}
Protocol~3 evaluates generalization from scarce data by training on only two source domains to test on the remaining two (\textbf{CW$\rightarrow$PS} and \textbf{PS$\rightarrow$CW}). This scenario is exceptionally challenging due to the reduced diversity of the training data.
Even with limited sources, RiSe consistently delivers the best performance. In the \textbf{CW$\rightarrow$PS} transfer, RiSe achieves a SoTA 8.05\% HTER and 96.63\% AUC, marking a significant lead over the next-best methods, MMDG++ (10.67\% HTER) and the top unimodal approach, FLIP (15.92\% HTER).
The reverse transfer, \textbf{PS$\rightarrow$CW}, is more difficult for all models due to a severe domain shift. While the performance margin tightens, RiSe still leads with 19.46\% HTER, narrowly outperforming DADM~\cite{dadm} (20.40\% HTER), whereas most other methods degrade to over 35-40\% HTER.
RiSe's advantage in this data-constrained setting stems from its ability to extract generalizable signals from minimal data. AsyIRM identifies domain-agnostic features common to the few source domains, while MMSD prevents overfitting to spurious modal biases, ensuring efficient use of all available sensor information. 
In summary, Protocol~3 demonstrates that RiSe is exceptionally robust when the distribution of training data is limited, maintaining a clear performance advantage even in the more difficult scenario.

\begin{table}[t]
    \caption{Coarse-grained ablation analysis on the proposed AsyIRM and MMSD. Average results on \textbf{CPS$\rightarrow$W}, \textbf{CPW$\rightarrow$S}, \textbf{CSW$\rightarrow$P}, and \textbf{PSW$\rightarrow$C} are reported.}
    \vspace{-0.8em}
    \renewcommand\arraystretch{1.2}
    \begin{tabularx}{\linewidth}{ccc| *{5}{>{\centering\arraybackslash}X}}
        \toprule
        \multirow{2}{*}{\textbf{Baseline}} & \multirow{2}{*}{{\begin{tabular}[c]{@{}c@{}}\textbf{AsyIRM}\\ \textit{\color{violet}(Risk 1 $\downarrow$)}\end{tabular}}} & \multirow{2}{*}{{\begin{tabular}[c]{@{}c@{}}\textbf{MMSD}\\ \textit{\color{teal}(Risk 2  $\downarrow$)}\end{tabular}}} & \multicolumn{2}{c}{\textbf{Average}} \\ \cmidrule(l){4-5} 
         &  &  & HTER (\%) $\downarrow$ & AUC (\%) $\uparrow$ \\ \midrule
        \ding{51} & - & - & 16.11 & 90.83 \\
        \ding{51} & \ding{51} & - & 11.38 & 94.17 \\
        \ding{51} & - & \ding{51} & 10.21 & 95.53 \\
        \rowcolor{lightblue!70}
        \ding{51} & \ding{51} & \ding{51} & \textbf{7.70} & \textbf{96.69} \\ \bottomrule
    \end{tabularx}
    \label{tab:abla}
\end{table}

\begin{table}[t]
    \caption{Fine-grained ablation analysis on the proposed AsyIRM. Average results on \textbf{CPS$\rightarrow$W}, \textbf{CPW$\rightarrow$S}, \textbf{CSW$\rightarrow$P}, and \textbf{PSW$\rightarrow$C} are reported.}
    \renewcommand\arraystretch{1.15}
    \vspace{-0.8em}
    \begin{tabularx}{\linewidth}{ccc| *{5}{>{\centering\arraybackslash}X}}
        \toprule
        \multirow{2}{*}{\textbf{\begin{tabular}[c]{@{}c@{}}IRM\\ Type\end{tabular}}} & \multirow{2}{*}{\textbf{\begin{tabular}[c]{@{}c@{}}Radial\\ Align\end{tabular}}} & \multirow{2}{*}{\textbf{\begin{tabular}[c]{@{}c@{}}Angular\\ Separate\end{tabular}}} & \multicolumn{2}{c}{\textbf{Average}} \\ \cmidrule(l){4-5} 
         &  &  & HTER (\%) $\downarrow$ & AUC (\%) $\uparrow$ \\ \midrule
        w/o IRM & - & - & 10.21 & 95.53 \\
        \midrule
        Vanilla IRM & \ding{51} & \ding{51} & 9.47 & 95.50 \\
        Rev. AsyIRM & \ding{51} & \ding{51} & 12.93 & 92.92 \\
        \rowcolor{lightblue!70}
        \textbf{AsyIRM} & \textbf{\ding{51}} & \textbf{\ding{51}} & \textbf{7.70} & \textbf{96.69} \\ \midrule
        AsyIRM & \ding{51} & - & 9.54 & 94.75 \\
        AsyIRM & - & \ding{51} & 9.58 & 95.30 \\
        \rowcolor{lightblue!70}
        \textbf{AsyIRM} & \textbf{\ding{51}} & \textbf{\ding{51}} & \textbf{7.70} & \textbf{96.69} \\ \bottomrule
    \end{tabularx}
    \label{tab:abla-asyirm}
\end{table}

\subsection{Ablation Study}
In this subsection, we first perform detailed ablation analysis on the key components of RiSe, i.e., AsyIRM and MMSD, to evaluate their contribution.
Then, we conduct a sensitive analysis on important hyperparameters.
In Table~\ref{tab:abla}, we present a coarse-grained ablation to evaluate the contributions of the AsyIRM and MMSD modules. Average results across four LOO sub-protocols in Protocol 1 are reported.

\subsubsection{Effectiveness of AsyIRM and MMSD}
To quantify the impact of our proposed modules, we evaluate four configurations: a baseline model, the baseline augmented with only AsyIRM, the baseline with only MMSD, and the complete RiSe model incorporating both. 
The results, summarized in Table~\ref{tab:abla}, reveal the complementary nature.
Integrating AsyIRM, which enforces a domain-invariant asymmetric decision boundary, yields a substantial performance gain, reducing the HTER to 11.38\%. 
Similarly, applying MMSD, which focuses on learning robust and disentangled features, reduces the HTER to 10.21\%. 
While both modules are individually effective, their combination results in the best performance, achieving an HTER of 7.70\% and an AUC of 96.69\%.
This represents an 8.41\% absolute reduction in HTER from the baseline, which corresponds to a 52\% relative reduction in error. 
This demonstrates a powerful module synergy: AsyIRM optimizes \textit{Risk~1}, while MMSD reduces \textit{Risk~2}, and addressing both aspects leads to a SoTA generalization capability.

\begin{figure*}
    \centering
    \includegraphics[width=\linewidth]{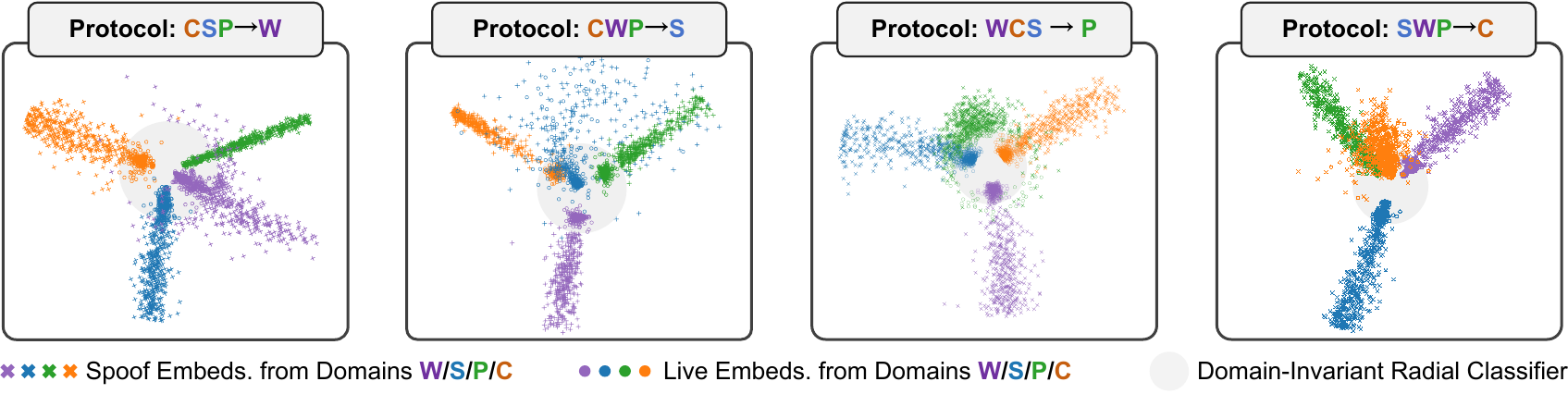}
    \caption{Visualization of features learned by AsyIRM across different LOO protocols. AsyIRM disentangles domain (encoded in angle) and liveness (encoded in norm) information in the embedding space. 
    }
    \label{fig:tsne}
\end{figure*}

\begin{table}[t]
    \caption{Fine-grained ablation analysis on the proposed MMSD. Average results on \textbf{CPS$\rightarrow$W}, \textbf{CPW$\rightarrow$S}, \textbf{CSW$\rightarrow$P}, and \textbf{PSW$\rightarrow$C} are reported.}
    \renewcommand\arraystretch{1.2}
    \vspace{-0.8em}
    \begin{tabularx}{\linewidth}{ccc| *{5}{>{\centering\arraybackslash}X}}
        \toprule
        \multirow{2}{*}{\textbf{\begin{tabular}[c]{@{}c@{}}Feature\\ Mixing\end{tabular}}} & \multirow{2}{*}{\textbf{\begin{tabular}[c]{@{}c@{}}Feature\\ Disentangle\end{tabular}}} & \multirow{2}{*}{\textbf{\begin{tabular}[c]{@{}c@{}}Spatial\\ Perturb\end{tabular}}} & \multicolumn{2}{c}{\textbf{Average}} \\ \cmidrule(l){4-5} 
         &  &  & HTER (\%) $\downarrow$ & AUC (\%) $\uparrow$ \\ \midrule
        - & - & - & 11.38 & 94.17 \\ \midrule
        - & - & \ding{51} & 10.04 & 95.34 \\
        S & Modality & \ding{51} & 9.49 & 95.50 \\
        F & HF & \ding{51} & 8.47 & 95.96 \\
        F & LF & \ding{51} & 8.74 & 95.65 \\
        \rowcolor{lightblue!70}
        \textbf{F} & \textbf{HF\&LF} & \ding{51} & \textbf{7.70} & \textbf{96.69} \\ \midrule
        S & Modality & - & 9.54 & 95.29 \\
        F & HF\&LF & - & 8.28 & 95.59 \\
        \rowcolor{lightblue!70}
        \textbf{F} & \textbf{HF\&LF} & \ding{51} & \textbf{7.70} & \textbf{96.69} \\ \bottomrule
    \end{tabularx}
    \par\vspace{1mm}
    \begin{minipage}{\linewidth}
    \footnotesize
    \noindent
    *Note: S = Spatial, F = Frequency, HF = High Frequency, LF = Low Frequency.
    \end{minipage}
    \label{tab:abla-mmsd}
\end{table}

\subsubsection{Fine-Grained Analysis on AsyIRM}
The core hypothesis of AsyIRM is that real faces form a compact manifold while spoof attacks are diversely scattered. 
As shown in Table~\ref{tab:abla-asyirm}, we provide a fine-grained analysis of several IRM-based classifier variants to justify the design choices behind AsyIRM. 
The \textit{Vanilla IRM} variant, which uses a conventional symmetric radial classifier for the two classes, achieves an HTER of 9.47\%.
The ``Reversed AsyIRM'' variant, which inverts this assumption by placing spoofs inside a hypersphere and real faces outside, suffers a catastrophic performance collapse, with HTER soaring to 12.93\%. 
This result is significantly worse than even the model without any IRM component (10.21\% HTER), confirming that an incorrect geometric prior is more detrimental than none at all. 
Furthermore, removing either the radial alignment or the angular separation component from the full AsyIRM model degrades performance substantially (to 9.58\% and 9.54\% HTER, respectively), validating that both elements are essential for establishing a tight, domain-invariant decision boundary.

To further illustrate AsyIRM's effectiveness, we visualize the learned 2D embeddings of different source domains for Protocol 1 in Fig.~\ref{fig:tsne}. 
Each subplot displays the features after applying our proposed AsyIRM. 
We observe a clear disentanglement: features from different domains (regardless of live/spoof label) are distinctly separated along different angular directions. 
Concurrently, live embeddings (circles) are consistently clustered inside the inner gray circle (smaller norm), while spoof embeddings (crosses) are pushed to the outer region (larger norm), extending beyond the gray radial classifier.
This visualization strongly supports our claim that AsyIRM successfully encodes domain information in the angular component and liveness information in the radial component, leading to a robust and disentangled embedding space where live/spoof classification is norm-based and domain separation is angle-based.

\subsubsection{Fine-Grained Analysis on MMSD}
Table~\ref{tab:abla-mmsd} reports an ablation study on the design of the proposed MMSD module, highlighting the effects of spatial feature mixing (denoted ``S''), frequency-based feature mixing (``F''), feature disentanglement strategies, and spatial perturbation. 
First, frequency-domain feature mixing is more effective for this task than spatial-domain mixing. The best frequency-based configuration (``F + HF\&LF'') achieves an HTER of 8.28\%, significantly outperforming the best spatial-mixing approach (``S + Modality'') at 9.49\%. 
The results indicate that spatial mixing introduces crude local artifacts that encourage shortcut learning. 
In contrast, frequency mixing's global and semantically conflicting perturbations appear more effective at forcing the model to learn intrinsic, disentangled unimodal representations.
Second, we observe that a powerful and non-obvious synergy exists between frequency mixing and spatial perturbation. 
Adding spatial perturbation to the frequency-mixing model provides a substantial performance boost, reducing HTER from 8.28\% to the final 7.70\%. In contrast, adding the same perturbation to the spatial-mixing model is ineffective (HTER increases from 9.49\% to 9.54\%). 
The results indicate that while frequency mixing disrupts spurious content-based correlations (e.g., ``RGB reflection + Depth flatness''), spatial shuffling dismantles structure-based ones (e.g., consistent facial geometry), jointly forcing a more robust representation.

\begin{figure}[t]
    \centering
    \includegraphics[width=1\linewidth]{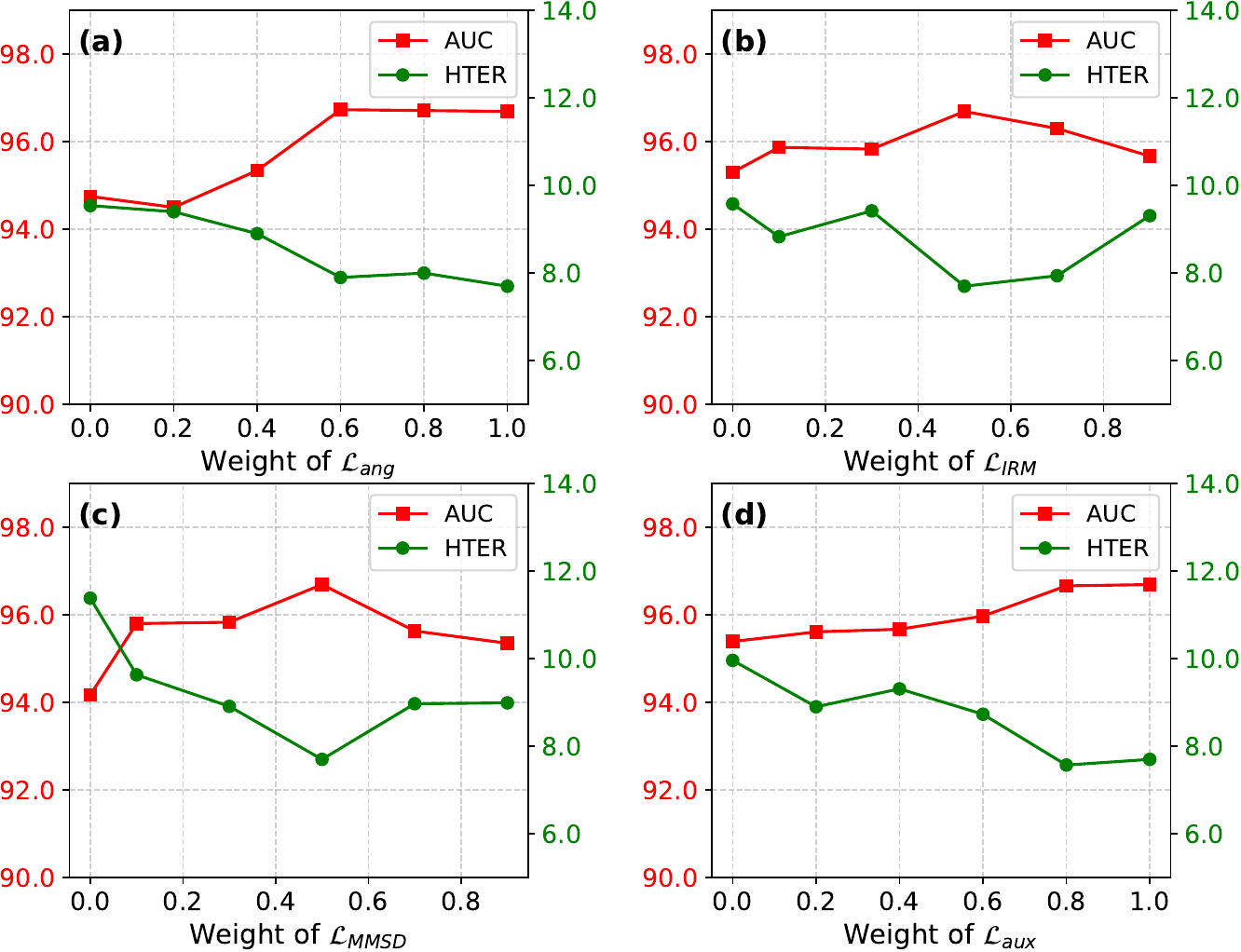}
    \vspace{-0.4em}
    \caption{Hyperparameter analysis of loss weights in Eq.~\ref{eq:loss-weight}. We evaluate the impact of the weights for (a) $\mathcal L_{\mathrm{ang}}$, (b) $\mathcal L_{\mathrm{IRM}}$, (c) $\mathcal L_{\mathrm{MMSD}}$, and (d) $\mathcal L_{\mathrm{aux}}$.}
    \label{fig:loss}
\end{figure}

\subsubsection{Hyperparameter Analysis}
Here, we analyze the sensitivity of RiSe to key hyperparameters, namely the weights assigned to different loss terms (i.e., $\lambda_{\mathrm{ang}}, \lambda_{\mathrm{IRM}}, \lambda_{\mathrm{MMSD}},\text{and}\,\lambda_{\mathrm{aux}}$) and initial radius $r$ of each radial classifier.

\vspace{1mm}
\noindent\textbf{Analysis on loss weights.}
As shown in Fig.~\ref{fig:loss}, our method's performance is highly robust to variations in the four loss weights ($\lambda_{\mathrm{ang}}$, $\lambda_{\mathrm{IRM}}$, $\lambda_{\mathrm{MMSD}}$, and $\lambda_{\mathrm{aux}}$) with the primary classification loss fixed at 1.0. Varying each hyperparameter yields only minor fluctuations in HTER and AUC. For example, even with $\lambda_{\mathrm{IRM}}$ set to an extreme value, HTER remains below 10\% and AUC stays above 95\%. 
Similarly, large changes in $\lambda_{\mathrm{ang}}$, $\lambda_{\mathrm{MMSD}}$, or $\lambda_{\mathrm{aux}}$ cause only slight performance variations. 
Across most of these trials, RiSe consistently maintains a low HTER (around 9\%--9.5\%) and high AUC (about 95.5\%--96\%), outperforming the prior SoTA (MMDG++~\cite{xun2025tpami}, HTER=9.98\%, AUC=95.21\%). 
These observations indicate that the proposed RiSe exhibits desirable insensitivity to loss weight choices.

\vspace{1mm}
\noindent\textbf{Analysis on initial classification radius $r$.} 
We initialize the classifier's radius parameter using the softplus function (implemented by PyTorch) , i.e., $r\!=\!\varphi(s)\!=\!\mathrm{softplus}(s)$ to ensure non-negativity. As shown in the results, the choice of initialization has a clear impact: $\varphi(0.0)$ achieves the best performance (HTER 7.70\%, AUC 96.69\%), while extreme initializations such as $\varphi(-2.0)$ or $\varphi(3.0)$ lead to noticeably degraded results (HTER $>$10\%, AUC $\leq$95.24\%). Although moderate settings (e.g., $\varphi(-1.0)$ or $\varphi(1.0)$) still yield competitive results, the performance fluctuations indicate that RiSe exhibits a certain degree of sensitivity to large shifts in the initial radius value.

\section{Conclusion} \label{sec:conclusion}
In conclusion, this paper presented RiSe, a novel multimodal face anti-spoofing framework explicitly designed to improve cross-domain generalization. 
Leveraging an asymmetric invariant risk minimization principle alongside a multimodal synergy mechanism, the proposed approach learns representations that remain robust across diverse domains and modalities. 
Underpinning this design, we derived theoretical generalization error bounds, making it the first FAS framework with provable cross-domain guarantees. 
Our extensive evaluation across multiple cross-domain multimodal protocols confirmed that AsyIRM$+$MMSD achieves SoTA performance, surpassing previous methods. 
These results validated the efficacy of RiSe and demonstrate RiSe's practical importance for robust face anti-spoofing in real-world deployment.

Beyond face anti-spoofing, the contributions of this work may have broader implications for multimodal learning. 
The principles of invariant learning and modality-specific synergy introduced here could benefit tasks such as multimodal anomaly detection, audio-visual understanding, and multi-sensor healthcare analytics.
Looking ahead, a promising direction is to integrate test-time adaptation techniques, enabling models to dynamically adjust to domain shifts and further enhance multimodal generalization.

\begin{figure}[t]
    \centering
    \includegraphics[width=1\linewidth]{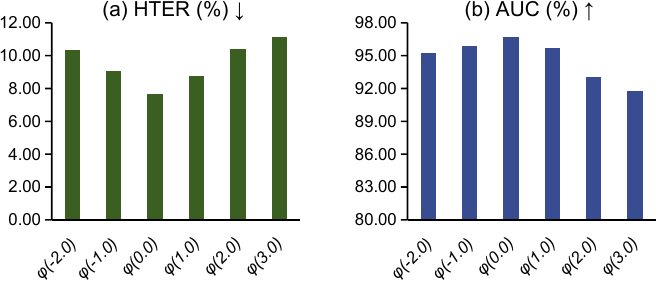}
    \vspace{-0.4em}
    \caption{Hyperparameter analysis on the initialization value $r\!=\!\varphi(s)$ for the radial classifier. Performance is measured in (a) HTER (lower is better) and (b) AUC (higher is better).}
    \label{fig:hyper}
\end{figure}

\bibliographystyle{IEEEtran}
\bibliography{IEEEabrv,reference}

\appendices
\section{Detailed Proof on Risk Decomposition}
\label{appendix:decomposition}
\begin{definition}[Jensen-Shannon (JS) Divergence~\cite{js}]
    The JS divergence~\cite{js} is a symmetrized and smoothed version of the KL divergence~\cite{kl}. For distributions $\mathcal P$ and $\mathcal Q$, it is defined as:
    \begin{equation}
        D_{\mathrm{JS}}(\mathcal P \| \mathcal Q) = \frac{1}{2} \mathrm{KL}(\mathcal P \| \mathcal J) + \frac{1}{2} \mathrm{KL}(Q \| \mathcal J),
    \end{equation}
    where $\mathcal J = \frac{1}{2}(\mathcal P + \mathcal Q)$ is the average distribution.
\end{definition}

\begin{lemma}[Triangle Inequality of $D_{\mathrm{JS}}$]
The square root of the JS divergence (JSD), $D_{\mathrm{JS}}(\cdot \| \cdot)$, is a metric on the space of probability distributions. Specifically, it satisfies the triangle inequality. 
For any three probability distributions $\mathcal P, \mathcal Q, \mathcal G$ defined on the same space $\mathcal{X}$, we have:
\begin{equation}
	\sqrt{D_{\mathrm{JS}}(\mathcal P \| \mathcal G)} \le \sqrt{D_{\mathrm{JS}}(\mathcal P \| \mathcal Q)} + \sqrt{D_{\mathrm{JS}}(\mathcal Q \| \mathcal G)}.
\end{equation}
\end{lemma}

\begin{proof}
The proof relies on a geometric interpretation of probability distributions. 
We can map any discrete probability distribution $\mathcal P = \{p_1, p_2, \dots, p_n\}$ to a vector on the positive orthant of a unit hypersphere in $\mathbb{R}^n$.

\vspace{2mm}
\noindent\textbf{Step 1: Mapping Distributions to a Hypersphere.}
Let us define a mapping $\psi: \mathcal P \to \boldsymbol{v}_{\mathcal P}$ where $\boldsymbol{v}_{\mathcal P} \in \mathbb{R}^n$ is a vector whose components are the square roots of the probabilities in $\mathcal P$:
\begin{equation}
	\boldsymbol{v}_{\mathcal P} = \left( \sqrt{p_1}, \sqrt{p_2}, \dots, \sqrt{p_n} \right).
\end{equation}

Since $\sum_{i=1}^n p_i = 1$, the squared L2-norm of this vector is $\|\boldsymbol{v}_{\mathcal P}\|_2^2 = \sum_{i=1}^n (\sqrt{p_i})^2 = \sum_{i=1}^n p_i = 1$. 
Thus, all such vectors lie on the surface of a unit hypersphere.

\vspace{2mm}
\noindent\textbf{Step 2: Relating JSD to the Hypersphere Geometry.}
The Euclidean distance between two such vectors, $\boldsymbol{v}_{\mathcal P}$ and $\boldsymbol{v}_{\mathcal Q}$, in this embedded space is a standard metric and inherently satisfies the triangle inequality. Let us analyze the squared Euclidean distance:
\begin{equation}
\label{eq:euclidean_dist}
\begin{aligned}
    \|\boldsymbol{v}_{\mathcal P} - \boldsymbol{v}_{\mathcal Q}\|_2^2 &= \sum_{i=1}^n (\sqrt{p_i} - \sqrt{q_i})^2 \\
    &= \sum_{i=1}^n (p_i - 2\sqrt{p_i q_i} + q_i) \\
    &= \sum p_i + \sum q_i - 2 \sum \sqrt{p_i q_i} \\
    &= 2 \left( 1 - \sum_{i=1}^n \sqrt{p_i q_i} \right).
\end{aligned}
\end{equation}

This distance is directly related to the Hellinger distance, defined as $H(\mathcal P,\mathcal Q) = \frac{1}{\sqrt{2}} \|\boldsymbol{v}_{\mathcal P} - \boldsymbol{v}_{\mathcal Q}\|_2$.
It is a well-known result, established by Endres and Schindelin (2003), that the Jensen-Shannon divergence is equal to the squared distance to the mean in this hypersphere embedding. 
More formally, for the average distribution $\mathcal J = \frac{1}{2}(\mathcal P+\mathcal Q)$, its corresponding vector is $\boldsymbol{v}_{\mathcal J} = (\sqrt{j_1}, \dots, \sqrt{j_n})$. The JSD can be expressed as:
\begin{equation}
\label{eq:jsd_geo}
    D_{\mathrm{JS}}(\mathcal P \| \mathcal Q) = \frac{1}{2} \|\boldsymbol{v}_{\mathcal P} - \boldsymbol{v}_{\mathcal J}\|_2^2 + \frac{1}{2} \|\boldsymbol{v}_{\mathcal Q} - \boldsymbol{v}_{\mathcal J}\|_2^2.
\end{equation}
Furthermore, it has been shown that $\sqrt{\text{JSD}(\mathcal P\|\mathcal Q)}$ corresponds to a metric distance on the manifold of probability distributions. While a full proof involving the Fisher information metric is more rigorous, a key insight is that $\sqrt{D_{\mathrm{JS}}(\cdot\|\cdot)}$ is monotonically related to the great-circle distance (geodesic) on the hypersphere between $\boldsymbol{v}_{\mathcal P}$ and $\boldsymbol{v}_{\mathcal Q}$.

\vspace{2mm}
\noindent\textbf{Step 3: Applying the Triangle Inequality.}
Since the vectors $\boldsymbol{v}_{\mathcal P}, \boldsymbol{v}_{\mathcal Q}, \boldsymbol{v}_{\mathcal J}$ are points in a Euclidean space (and on a hypersphere), their Euclidean distances must satisfy the triangle inequality:
\begin{equation}
	\|\boldsymbol{v}_{\mathcal P} - \boldsymbol{v}_{\mathcal J}\|_2 \le \|\boldsymbol{v}_{\mathcal P} - \boldsymbol{v}_{\mathcal Q}\|_2 + \|\boldsymbol{v}_{\mathcal Q} - \boldsymbol{v}_\mathcal{J}\|_2.
\end{equation}

Because $\sqrt{D_{\mathrm{JS}}(\cdot\|\cdot)}$ is a metric distance on the statistical manifold (a property that can be shown to be equivalent to being a distance in the hypersphere representation), it must also satisfy the triangle inequality. 
The relationship between $\sqrt{D_{\mathrm{JS}}(\cdot\|\cdot)}$ and distances on the sphere is non-trivial but established. Therefore, by leveraging this known property, we can state that for any three distributions $\mathcal P, \mathcal Q, \mathcal G$:
\begin{equation}
    \sqrt{D_{\mathrm{JS}}(\mathcal P \| \mathcal G)} \le \sqrt{D_{\mathrm{JS}}(\mathcal P \| \mathcal Q)} + \sqrt{D_{\mathrm{JS}}(\mathcal Q \| \mathcal G)}.
\end{equation}

\end{proof}

\section{Derivation on Asymmetric Distributions}
\subsection{Detailed Derivation of Assumption~\ref{assum:unimodal_revised}}
\label{appendix:unimodal_revised}
The spoof class consists of a diverse set of attack types with heterogeneous physical properties and media. 
Importantly, the spoof class itself does not possess a shared intrinsic essence; it merely represents the complement of the live class. 
Each attack method introduces its own distinctive physical artifacts that differ from genuine facial characteristics. 

During training, the encoder $\phi(\cdot)$ learns to capture these discrepancies. 
We decompose the embedding of a spoof sample as
\begin{equation}
\boldsymbol z_{\mathrm{spoof}} = \boldsymbol z_{\mathrm{face}} + \boldsymbol \delta_{\mathrm{artifact}},
\end{equation}
where $\boldsymbol z_{\mathrm{face}}$ represents the embedding corresponding to the facial appearance, and $\boldsymbol \delta_{\mathrm{artifact}}$ denotes the feature vector specifically encoding spoof-related artifacts. 
Our goal is to analyze the distribution of $\boldsymbol \delta_{\mathrm{artifact}}$. 
Since artifacts from different attack types are physically distinct, the corresponding $\boldsymbol\delta_{\mathrm{artifact}}$ vectors are approximately orthogonal (or at least widely separated in angle). 

Suppose the dataset contains $K$ attack types. The spoof class ($y\!=\!1$) in the embedding space can then be modeled as a Gaussian mixture:
\begin{equation}
\begin{aligned}
    p(\boldsymbol z \mid y=1) &= \sum_{k=1}^{K} \pi_k \, p(\boldsymbol z\mid \text{attack}_k), \\
    p(\boldsymbol z \mid \text{attack}_k) &= \mathcal N(\boldsymbol z; \mu_k, \Sigma_k),
\end{aligned}
\end{equation}
where $\pi_k$ is the prior of attack $k$, and $\mu_k \approx \delta_k$ is its unique artifact direction. 
To align with the isotropic, small-variance distribution of genuine samples ($\boldsymbol z\mid y\!=\!0 \sim \mathcal N(0,\sigma_0^2 I_d)$), we impose the following uniform isotropic assumption on intra-class variance across attacks:
\begin{equation}
    \ \Sigma_k = \sigma_{\mathrm{eff}}^2 I_d
\quad \forall k, \qquad
\sigma_{\mathrm{eff}}^2 \triangleq \sigma_0^2 + \varepsilon,\;\; 0<\varepsilon\ll\sigma_0^2.\ 
\end{equation}
That is, spoof embeddings share the baseline isotropic noise $\sigma_0^2 I_d$ of live samples, plus a small isotropic inflation $\varepsilon I_d$ within each attack subspace. 
For simplicity, we denote the common intra-class variance by $\sigma_{\mathrm{eff}}^2$. 

\noindent\textbf{Second moment and equivalent decomposition.}  
By linearity of mixtures and the Gaussian second-moment identity $\mathbb E[\boldsymbol z \boldsymbol z^\top]=\Sigma+\mu\mu^\top$, we obtain
\begin{equation}
\small
\mathbb{E}[\, \boldsymbol z \boldsymbol z^\top \mid y\!=\!1 \,]
= \sum_{k=1}^{K} \pi_k \big( \Sigma_k + \mu_k \mu_k^\top \big)
= \sigma_{\mathrm{eff}}^2 I_d + \sum_{k=1}^{K} \pi_k\,\mu_k\mu_k^\top.
\end{equation}

\noindent\textbf{Spectral structure (orthogonal or near-orthogonal case).}  
This yields a \emph{spiked covariance structure}: an isotropic noise floor plus a low-rank signal term. 
The principal components align with $\mathrm{span}\{\mu_k\}$, while directions orthogonal to this span retain only isotropic noise $\sigma_{\mathrm{eff}}^2$. 
If we further approximate $\{\mu_k\}$ as mutually orthogonal, then each $\mu_k$ is an eigenvector with eigenvalue
\begin{equation}
    \lambda_k = \sigma_{\mathrm{eff}}^2 + \pi_k\|\mu_k\|^2,
\end{equation}
while all other orthogonal directions have eigenvalue $\sigma_{\mathrm{eff}}^2$. 
In the near-orthogonal case, these principal components and eigenvalues can be approximated stably via the spectrum of the Gram matrix
\(
G\!=\![\sqrt{\pi_i\pi_j}\,\mu_i^\top\mu_j]_{i,j},
\)
using perturbation results such as Davis–Kahan~\cite{davis_kahan}.

\subsection{Detailed Derivation of Assumption~\ref{assumption:multimodal_heterogeneous_v2}}
\label{appendix:multimodal_heterogeneous}
Here, we extend the above unimodal derivation to the multimodal setting. 
For a multimodal input, the embedding $\boldsymbol z \in \mathbb{R}^{M\times d}$ is the concatenation of $M$ modality-specific feature vectors, $\boldsymbol z = [\boldsymbol z_1^\top, \dots, \boldsymbol z_M^\top]^\top$, where each $\boldsymbol z_m \in \mathbb{R}^d$.
Similar to the unimodal case, we can decompose the embedding of a spoof sample from modality $m$ as:
\begin{equation*}
    \boldsymbol z_m = \boldsymbol z_{m, \text{face}} + \boldsymbol \delta_{m, \text{artifact}}.
\end{equation*}
Here, $\boldsymbol z_{m, \text{face}}$ represents the facial appearance as captured by modality $m$, and $ \delta_{m, \text{artifact}}$ is the feature vector encoding the spoof artifacts specific to that modality. The full multimodal embedding is the concatenation of these vectors.

The spoof class is a mixture over $K$ distinct attack types. For a specific attack type $k$, the embedding can be modeled as a multivariate Gaussian. The key difference in the multimodal setting is that the noise is no longer uniform across all feature dimensions. Each modality introduces its own sensor noise and is affected by artifacts differently.
We model the distribution for attack type $k$ as:
\begin{equation}
    p(\boldsymbol z \mid \text{attack}_k) = \mathcal{N}(\boldsymbol z;  \mu_k,  \Sigma_k),
\end{equation}
where $ \mu_k = [\mu_{k,1}^\top, \dots, \mu_{k,M}^\top]^\top$ is the concatenated mean vector representing the unique artifact signature of attack $k$ across all $M$ modalities.

The crucial extension lies in the covariance matrix $\boldsymbol \Sigma_k$. In the unimodal case, we assumed a simple isotropic covariance $\sigma_{\text{eff}}^2 I_d$. In the multimodal case, we assume that the noise across different modalities is uncorrelated, but the noise level within each modality is different. 
This naturally leads to a block-diagonal covariance structure.

Consistent with the unimodal assumption, we model the intra-attack variance for each modality $m$ as being isotropic, but with a modality-specific effective variance $\sigma_{\text{eff}, m}^2$. This variance accounts for both the baseline noise of live samples in that modality and a small inflation due to the attack.
\begin{equation}
    \Sigma_k = \mathrm{diag}(\sigma_{\text{eff},1}^2 I_d, \sigma_{\text{eff},2}^2 I_d, \dots, \sigma_{\text{eff},M}^2 I_d) \triangleq \sigma_{\mathrm{eff,multi}}.
\end{equation}
Importantly, we assume this effective noise floor $\boldsymbol \sigma_{\mathrm{eff,multi}}$ is common across all attack types $k$, as it primarily reflects sensor characteristics rather than specific attack properties.

The distribution of the entire spoof class ($y=1$) is a mixture of these attack-specific Gaussians:
\begin{equation}
    p(\boldsymbol z \mid y=1) = \sum_{k=1}^{K} \pi_k \, \mathcal{N}(\boldsymbol z; \mu_k,  \sigma_{\mathrm{eff,multi}}),
\end{equation}
where $\pi_k$ is the prior probability of attack type $k$.

Using the linearity of expectation and the second-moment identity for Gaussians, we can derive the second-moment matrix for the multimodal spoof class:
\begin{equation}
\label{eq:second_moment_multi}
\begin{aligned}
    \mathbb{E}[\boldsymbol z \boldsymbol z^\top \mid y=1] &= \sum_{k=1}^{K} \pi_k \, \mathbb{E}_{\boldsymbol z \sim \mathcal{N}(\mu_k, \sigma_{\mathrm{eff,multi}})}[\boldsymbol z \boldsymbol z^\top] \\
    &= \sum_{k=1}^{K} \pi_k ( \sigma_{\mathrm{eff,multi}} +  \mu_k  \mu_k^\top) \\
    &= \Big(\sum_{k=1}^{K} \pi_k\Big)  \sigma_{\mathrm{eff,multi}} + \sum_{k=1}^{K} \pi_k  \mu_k  \mu_k^\top \\
    &=  \sigma_{\mathrm{eff,multi}} + \sum_{k=1}^{K} \pi_k  \mu_k  \mu_k^\top.
\end{aligned}
\end{equation}

This derivation formally shows how multiple modalities amplify the distributional asymmetry between the Live and Spoof classes:

\begin{enumerate}
    \item \textbf{Higher-Dimensional Support}: The spoof embeddings now live in a higher-dimensional space ($\mathbb{R}^{M\times d}$ vs. $\mathbb{R}^d$), providing more dimensions for discrepancies to manifest.

    \item \textbf{Heterogeneous Noise Floor}: Unlike the simple isotropic noise floor $\sigma_{\text{eff}}^2 I_d$ in the unimodal case, the multimodal spoof distribution has a more complex, block-diagonal noise floor $\boldsymbol \sigma_{\mathrm{eff,multi}}$. The eigenvalues of this covariance matrix are no longer a single value $\sigma_{\text{eff}}^2$, but a set of values $\{\sigma_{\text{eff},1}^2, \dots, \sigma_{\text{eff},M}^2\}$.
    
    \item \textbf{Spiked Covariance on a Heterogeneous Base}: The overall second-moment matrix retains the ``spiked covariance'' structure, but the ``spikes'' (from the low-rank term $\sum \pi_k  \mu_k  \mu_k^\top$) are now added to a non-uniform, heterogeneous base.
\end{enumerate}

In contrast, the Live class is typically modeled as a simple, compact distribution around the origin, often as a single isotropic Gaussian $\mathcal{N}(\boldsymbol 0, \boldsymbol \sigma_{0, \text{multi}})$, where $\boldsymbol \sigma_{0, \text{multi}}$ might also be block-diagonal but with smaller, more uniform variances. The introduction of modality-specific variances and the expansion into a higher-dimensional space makes the spoof distribution's shape significantly more complex and spread out than the live distribution, thus \textbf{amplifying the distributional gap} between them.

\section{Detailed Proof on IRM vs. AsyIRM}
\begin{lemma}[Negative Log-Posterior Probability]
Define $\mathcal L(\cdot)$ as the negative log-posterior probability:
\[
\mathcal L(h) = - \ln \mathcal P(h\,|\,S),
\]
where $\mathcal P(h\,|\,S)$ denotes the posterior probability of a hypothesis (or model parameter) $h$ given the observed training data $S$.
\end{lemma}

\vspace{2mm}
\begin{lemma}[KL Divergence Between Multivariate Gaussian Distributions]
    \label{lemma:kl}
    The KL divergence between two multivariate Gaussian distributions $\mathcal N_1(\mu_1, \Sigma_1)$ and $\mathcal N_2(\mu_2, \Sigma_2)$ is given by:
    \begin{equation}
    \footnotesize
        \label{eq:kl}
        \mathrm{KL} (\mathcal N_1\,||\,\mathcal N_2) = \frac{1}{2}\Bigl(\mathrm{tr}(\Sigma_2^{-1}\Sigma_1) + (\mu_2-\mu_1)^\top\Sigma_2^{-1}(\mu_2-\mu_1)-d+\ln\tfrac{\mathrm{det} \Sigma_2}{\mathrm{det} \Sigma_1}\Bigr),  
    \end{equation}
    where $d$ is the dimensionality of the distributions.
\end{lemma}
\subsection{Proof of Proposition~\ref{proposition:sym_kl}}
\label{appendix:sym_kl}
\begin{proof}
Consider a hyperplane classifier $\mathcal F_\beta(z)=\beta^\top z$ (bias term omitted for clarity, but can be added via concatenating a constant feature).  
With the logistic model ($y\in\{0(\mathrm{live}),1(\mathrm{spoof})\}$):
\begin{equation}
	\begin{aligned}
		P(y\mid z,\beta)&=\sigma(y\,\beta^\top z), \\
		\sigma(t)&=\frac{1}{1+e^{-t}}.
	\end{aligned}
\end{equation}

\noindent\textbf{Likelihood.}
For an i.i.d. dataset $\mathcal D=\{(z_i,y_i)\}_{i=1}^N$,  
\begin{equation}
	\begin{aligned}
		P(\mathcal D\mid \beta)&=\prod_{i=1}^N\sigma(y_i\,\beta^\top z_i), \\
		\ln P(\mathcal D\mid \beta)&=\sum_{i=1}^N \ln \sigma(y_i\,\beta^\top z_i).
	\end{aligned}
\end{equation}

\noindent\textbf{Prior.}
An isotropic Gaussian prior is imposed:
\begin{equation}
	\begin{aligned}
		\Pi_{\mathrm{sym}}(\beta)&=\mathcal N(\beta; 0,\sigma_\beta^2 I_d),\\
		\ln \Pi_{\mathrm{sym}}(\beta)&=-\frac{1}{2\sigma_\beta^2}\|\beta\|_2^2 - \frac{d}{2}\ln(2\pi\sigma_\beta^2).
	\end{aligned}
\end{equation}

\noindent\textbf{Posterior.}
By Bayes's rule, $P(\beta\mid \mathcal S)\propto P(\mathcal S \mid \beta)\,\Pi_{\mathrm{sym}}(\beta)$.  
The negative log-posterior (up to constants $C_0$) is:
\begin{equation}
	\begin{aligned}
		\mathcal L(\beta)&\triangleq-\ln P(\beta\mid \mathcal D)\\
		&=\sum_{i=1}^N\ln(1+e^{-y_i\,\beta^\top z_i})+\frac{1}{2\sigma_\beta^2}\|\beta\|_2^2+C_0.		
	\end{aligned}
\end{equation}

At the Maximum A Posteriori (MAP) solution $\beta_{\mathrm{MAP}}=\arg\min_\beta \mathcal L(\beta)$, the Laplace approximation yields 
\(\mathcal Q_{\mathrm{sym}}(\beta)=\mathcal N(\beta_{\mathrm{MAP}},H_{\mathrm{sym}}^{-1})\), 
where
\begin{equation}
	\begin{aligned}
		H_{\mathrm{sym}}&=\sum_{i=1}^N c_i z_i z_i^\top + \frac{1}{\sigma_\beta^2}I_d,\\
		c_i&=\sigma(a_i)\Big(1-\sigma(a_i)\Big)\in(0,\frac{1}{4}],\\
		a_i&=y_i\beta_{\mathrm{MAP}}^\top z_i.
	\end{aligned}
\end{equation}
\noindent\textbf{Expected structure.}
Let $\bar c=\mathbb E[c_i]$ and adopt Assumption~\ref{assum:unimodal_revised}:
\begin{equation}
	\begin{aligned}
		p(\boldsymbol z\mid y=0) &\sim \mathcal N(0,\sigma_0^2 I_d),\\
		p(\boldsymbol z\mid y=1)&=\sum_{k=1}^K \pi_k\,\mathcal N(\mu_k,\sigma_{\mathrm{eff}}^2 I_d),		
	\end{aligned}
\end{equation}
with $\sigma_{\mathrm{eff}}^2=\sigma_0^2+\varepsilon$.  
Defining class priors:
\begin{equation}
	\begin{aligned}
		\Pi_0&=\mathbb P(y=0),\quad\Pi_1=\mathbb P(y=1),\\
		U&=\sqrt{N\bar c\,\Pi_1}\,\Big[\sqrt{\pi_1}\mu_1,\ldots,\sqrt{\pi_K}\mu_K\Big]\in\mathbb R^{m\times K}, \\
		G&=U^\top U,\quad \tau=\mathrm{rank}(G)\le K,						
	\end{aligned}
\end{equation}
we can obtain:
\begin{equation}
	\begin{aligned}
		H_{\mathrm{sym}} &\approx \lambda_0 I_m + UU^\top, \\
		\lambda_0&=N\cdot\bar c\cdot(\Pi_0\sigma_0^2+\Pi_1\sigma_{\mathrm{eff}}^2)+\frac{1}{\sigma_\beta^2}.	
	\end{aligned}
\end{equation}

\noindent\textbf{Trace and determinant.}
By matrix identities,
\begin{equation}
	\begin{aligned}
		\operatorname{tr}(H_{\mathrm{sym}}^{-1})
		&= \frac{d}{\lambda_0} - \frac{1}{\lambda_0}\operatorname{tr}\!\Big(G(\lambda_0 I_K+G)^{-1}\Big) \\
		&= \frac{d-\tau}{\lambda_0}+\sum_{i=1}^\tau\frac{1}{\lambda_0+\lambda_i(G)}, \\
		\ln\det H_{\mathrm{sym}}
		&= d\ln\lambda_0+\ln\det\Big(I_K+\frac{1}{\lambda_0}G\Big)\\
		&= (d-\tau)\ln\lambda_0+\sum_{i=1}^\tau\ln\Big(\lambda_0+\lambda_i(G)\Big).
	\end{aligned}
\end{equation}

\noindent\textbf{KL divergence.}
From Definition~\ref{lemma:kl}, we have
\begin{equation}
\small
\begin{aligned}
	\mathrm{KL}(\mathcal Q_{\mathrm{sym}}&\Vert \Pi_{\mathrm{sym}})=\\
	&\frac{1}{2}\Big(
	\frac{\operatorname{tr}(H_{\mathrm{sym}}^{-1})+\|\beta_{\mathrm{MAP}}\|_2^2}{\sigma_\beta^2}
	- d + d\ln\sigma_\beta^2 + \ln\det H_{\mathrm{sym}}
	\Big).
\end{aligned}
\end{equation}

Let $t=\sigma_\beta^2\lambda_0\ge 1$ and $\gamma_i=\lambda_i(G)/\lambda_0\ge 0$. 
Substituting the above,
\begin{equation}
	\begin{aligned}
		\mathrm{KL}(\mathcal Q_{\mathrm{sym}}\Vert \Pi_{\mathrm{sym}}) &= \frac{d}{2}\Big(\frac{1}{t}+\ln t-1\Big)\\
		&+\frac{1}{2}\sum_{i=1}^\tau\Big[\ln(1+\gamma_i)-\frac{1}{t}\frac{\gamma_i}{1+\gamma_i}\Big] \\
		&+\frac{\|\beta_{\mathrm{MAP}}\|_2^2}{2\sigma_\beta^2}.
	\end{aligned}
\end{equation}

Assuming $\|\beta_{\mathrm{MAP}}\|_2^2=O(d)$ (common under $L_2$ regularization), we conclude
\begin{equation}
	\mathrm{KL}(\mathcal Q_{\mathrm{sym}}\Vert \Pi_{\mathrm{sym}})=O(d)+O(K),
\end{equation}
\end{proof}

\subsection{Proof of Proposition~\ref{proposition:multi_kl}}
\label{appendix:multi_kl}
\begin{proof}
    The feature vector $\boldsymbol z \in \mathbb{R}^{M\times d}$ is a concatenation of $M$ modality-specific embeddings, each with dimension $d$. We write this as $\boldsymbol z = [\boldsymbol z_1^\top, \dots, \boldsymbol z_M^\top]^\top$.
    
    We consider a linear classifier $\mathcal{F}_{\boldsymbol\beta}(\boldsymbol z) = \boldsymbol\beta^\top \boldsymbol z$, where the weight vector $\boldsymbol\beta \in \mathbb{R}^{Md}$ is also a concatenation, $\boldsymbol\beta = [\boldsymbol\beta_1^\top, \dots, \boldsymbol\beta_M^\top]^\top$.

In the large sample limit, the Hessian matrix $H_{\mathrm{sym}}$ can be approximated as:
\begin{equation}
	\small
	H_{\mathrm{sym}} \approx N\bar{c} \left( \Pi_0 \mathbb{E}[\boldsymbol z \boldsymbol z^\top \mid y=0] + \Pi_1 \mathbb{E}[\boldsymbol z \boldsymbol z^\top \mid y=1] \right) + \frac{1}{\sigma_\beta^2} I_{Md}.
\end{equation}

Substituting the multimodal distributions, we get:
\begin{equation}
\small
\begin{aligned}
    H_{\mathrm{sym}} &\approx N\cdot\bar{c}\cdot\Big( \Pi_0 \sigma_{0,\text{multi}} + \Pi_1 \big( \sigma_{\mathrm{eff,multi}} + \sum_{k=1}^{K} \pi_k \mu_k \mu_k^\top \big) \Big) \\
    & + \frac{1}{\sigma_\beta^2} I_{Md} \\
    & = N\cdot\bar{c}\cdot(\Pi_0 \sigma_{0,\text{multi}} + \Pi_1 \sigma_{\mathrm{eff,multi}}) + \frac{1}{\sigma_\beta^2} I_{Md} \\
    &+ N\cdot\bar{c}\cdot\Pi_1 \sum_{k=1}^{K} \pi_k \mu_k \mu_k^\top.
\end{aligned}
\end{equation}
We decompose the Hessian into a block-diagonal base matrix $\boldsymbol\Lambda_0$ and a low-rank ``spike'' term $UU^\top$:
\begin{equation}
    H_{\mathrm{sym}} = \boldsymbol\Lambda_0 + UU^\top,
\end{equation}
where the spike term $U = \sqrt{N\cdot\bar{c}\cdot\Pi_1} [\sqrt{\pi_1}\mu_1, \dots, \sqrt{\pi_K}\mu_K] \in \mathbb{R}^{M\times d \times K}$; the block-diagonal base $\boldsymbol\Lambda_0 = \operatorname{diag}(\lambda_{0,1} I_d, \lambda_{0,2} I_d, \dots, \lambda_{0,M} I_d)$, with each block's eigenvalue being
    \begin{equation}
        \lambda_{0,m} = N\bar{c}(\Pi_0 \sigma_{0,m}^2 + \Pi_1 \sigma_{\mathrm{eff},m}^2) + \frac{1}{\sigma_\beta^2}.
    \end{equation}

The KL-divergence is given by:
\begin{equation}
	\begin{aligned}
  		\mathrm{KL}(\mathcal{Q}_{\mathrm{sym}} \| \Pi_{\mathrm{sym}}) &= \frac{1}{2} \Big( \frac{\mathrm{tr}(H_{\mathrm{sym}}^{-1}) + \|\boldsymbol\beta_{\mathrm{MAP}}\|_2^2}{\sigma_\beta^2} \\
  		&- Md + \ln\det(H_{\mathrm{sym}}) + Md\ln\sigma_\beta^2 \Big).		
	\end{aligned}
\end{equation}
Using the matrix determinant lemma, $\det(A+UV^\top) = \det(I+V^\top A^{-1}U)\det(A)$, we find the log-determinant:
\begin{equation}
\begin{aligned}
    \ln\det(H_{\mathrm{sym}}) &= \ln\det(\boldsymbol\Lambda_0 + UU^\top) \\
    &= \ln\det(\boldsymbol\Lambda_0) + \ln\det(I_K + U^\top \boldsymbol\Lambda_0^{-1} U) \\
    &= \sum_{m=1}^{M} d \ln(\lambda_{0,m}) + \ln\det(I_K + G_{\text{multi}}),
\end{aligned}
\end{equation}
where the generalized Gram matrix $G_{\text{multi}} = U^\top \boldsymbol\Lambda_0^{-1} U \in \mathbb{R}^{K \times K}$. Its $(i,j)$-th element is:
\begin{equation}
	\begin{aligned}
		(G_{\text{multi}})_{ij} &= \sqrt{\pi_i \pi_j} (N\cdot\bar{c}\cdot\Pi_1) \mu_i^\top \boldsymbol\Lambda_0^{-1} \mu_j \\
		&= \sqrt{\pi_i \pi_j} (N\cdot\bar{c}\cdot\Pi_1) \sum_{m=1}^{M} \frac{1}{\lambda_{0,m}} \mu_{i,m}^\top \mu_{j,m}.
	\end{aligned}
\end{equation}

Assuming $\|\boldsymbol\beta_{\mathrm{MAP}}\|_2^2 = O(M\times d)$, we can analyze the scaling of the dominant terms in the KL-divergence.

\begin{enumerate}
    \item \textbf{Dimension-dependent Term:} This term arises primarily from the log-determinant of the base matrix $\boldsymbol\Lambda_0$.
    \begin{equation}
    	\small
    	\begin{aligned}
    		\frac{1}{2} \left( \ln\det(\boldsymbol\Lambda_0) - Md \right) &= \frac{1}{2} \left( \sum_{m=1}^{M} d \ln(\lambda_{0,m}) - Md \right)\\ 
    		&= \frac{d}{2} \sum_{m=1}^{M} \Big(\ln(\lambda_{0,m}) - 1\Big) = O(M \cdot d).
    	\end{aligned}
    \end{equation}
    
    This term scales linearly with the total feature dimension, $M\cdot d$.

    \item \textbf{Attack-dependent Term:} This term arises from the low-rank spike structure, captured by $\ln\det(I_K + G_{\text{multi}})$.
    \begin{equation}
        \frac{1}{2} \ln\det(I_K + G_{\text{multi}}) = \frac{1}{2} \sum_{i=1}^{\mathrm{rank}(G)} \ln(1 + \lambda_i(G_{\text{multi}})).
    \end{equation}
    The heterogeneity of the noise variances across modalities ($\sigma_{\mathrm{eff},m}^2$) introduces a dependency on $M$ into the spectrum of the Gram matrix. 
\end{enumerate}

Combining the dominant terms, the overall scaling of the KL-divergence in the multimodal setting, under the assumption that $K$ is small relative to the total dimension, is:
\begin{equation}
    \mathrm{KL}(\mathcal{Q}_{\mathrm{sym}} \| \Pi_{\mathrm{sym}}) = O(d\cdot M) + O(K \log M).
\end{equation}

\end{proof}

\subsection{Proof of Proposition~\ref{proposition:asym_kl}}
\label{appendix:asym_kl}
\begin{proof}
Under Assumption~\ref{assum:unimodal_revised} (real faces isotropic with small variance; spoof samples as a Gaussian mixture with isotropic intra-class covariance $\Sigma_k=\sigma_{\mathrm{eff}}^2 I$), consider the asymmetric IRM classifier:
\begin{equation}
	\mathcal F_{r}(\boldsymbol z)=\|\boldsymbol z\|_2^2 - r,
\end{equation}
where we reparameterize the decision radius as a scalar $R=r^2$ for analytical convenience.  
The logistic likelihood is defined as:
\begin{equation}
	\begin{aligned}
		P(y\mid \boldsymbol z,R)&=\sigma\Big(y(\|\boldsymbol z\|^2-R)\Big),\\
		\Pi_{\mathrm{asym}}(R)&=\mathcal N(R; \mu_R,\sigma_R^2).
	\end{aligned}
\end{equation}

Thus the posterior $\mathcal P_{\text{asym}}(R)\propto P(S\mid R)\,\Pi_{\text{asym}}(R)$ yields a KL divergence:
\begin{equation}
	\mathrm{KL}\!\left(\mathcal P_{\text{asym}} \,\Vert\, \Pi_{\text{asym}}\right).
\end{equation}

\noindent\textbf{Negative log-posterior derivation.}
For $a_i(R)=y_i(\|\boldsymbol z_i\|^2-R)$, the single-sample negative log-likelihood (NLL) is $\ell_i(R)=\ln(1+e^{-a_i(R)})$.  
The overall negative log-posterior is
\begin{equation}
\mathcal L(R)=-\ln P(R\mid S)
=\sum_{i=1}^{N}\ell_i(R)+\frac{(R-\mu_R)^2}{2\sigma_R^2}+C_2.
\end{equation}
The derivatives are:
\begin{equation}
	\begin{aligned}
		\frac{\partial \ell_i}{\partial R}
		=\;&\sigma\big(-a_i(R)\big)\,y_i,\\
		\frac{\partial^2 \ell_i}{\partial R^2}
		=\;&\sigma\big(a_i(R)\big)\big(1-\sigma(a_i(R))\big) \;\triangleq\; c_i(R)\in\big(0,\frac{1}{4}\big].
	\end{aligned}
\end{equation}
Hence, we have:
\begin{equation}
	\begin{aligned}
		\nabla_R \mathcal L(R)&=\sum_{i=1}^{N} y_i\sigma\big(-a_i(R)\big)+\frac{R-\mu_R}{\sigma_R^2},\\
		H_{\text{asym}}(R)&=\sum_{i=1}^{N} c_i(R)+\frac{1}{\sigma_R^2}.
	\end{aligned}
\end{equation}
At $R_{\mathrm{MAP}}=\arg\min_R \mathcal L(R)$, the Laplace approximation gives:
\begin{equation}
	\begin{aligned}
		\mathcal Q_{\text{asym}}(R)&=\mathcal N\big(R\mid R_{\mathrm{MAP}},\,H_{\text{asym}}^{-1}\big),\\
		H_{\text{asym}}&=\sum_{i=1}^{N} c_i(R_{\mathrm{MAP}})+\frac{1}{\sigma_R^2}.
	\end{aligned}
\end{equation}
Since $c_i\in(0,\tfrac14]$ are bounded and independent of $m$, it follows that $H_{\text{asym}}=\Theta(N)+1/\sigma_R^2$.

\noindent\textbf{Closed-form KL and scaling.}
Using the 1D Gaussian KL formula:
\begin{equation}
	\begin{aligned}
		\mathrm{KL}&\Big(\mathcal Q_{\text{asym}} \,\Vert\, \Pi_{\text{asym}}\Big) =\\
		&\frac12\Big(
		\frac{H_{\text{asym}}^{-1}}{\sigma_R^2} +\frac{(R_{\mathrm{MAP}}-\mu_R)^2}{\sigma_R^2}
		 -1+\ln(\sigma_R^2 H_{\text{asym}})
		\Big).
	\end{aligned}
\end{equation}
In the large-sample limit, with $\bar c=\mathbb E\big[c_i(R_{\mathrm{MAP}})\big]\in(0,\tfrac14]$,
\begin{equation}
H_{\text{asym}}\ \approx\ N\cdot\bar c+\frac{1}{\sigma_R^2}.
\end{equation}
Thus,
\begin{equation}
	\begin{aligned}
		\mathrm{KL}\Big(\mathcal Q_{\text{asym}} \,\Vert\, \Pi_{\text{asym}}\Big)
		&=\frac12\Big(
		\frac{1}{\sigma_R^2(N\bar c+1/\sigma_R^2)}\\
		&+\frac{(R_{\mathrm{MAP}}-\mu_R)^2}{\sigma_R^2}
		-1\\
		&+\ln\big(\sigma_R^2(N\bar c+1/\sigma_R^2)\big)
		\Big),
	\end{aligned}
\end{equation}
which depends only on $N,\sigma_R^2,\bar c$, and the scalar offset $R_{\mathrm{MAP}}-\mu_R$, but is \textbf{independent of $d$ and $K$}.  
Therefore,
\begin{equation}
\mathrm{KL}\left(\mathcal Q_{\text{asym}} \,\Vert\, \Pi_{\text{asym}}\right)=O(1)\ \text{in $d,K$}.
\end{equation}

\noindent\textbf{Consistency with $r$-parameterization.}
If $r$ is used directly (instead of $R=r^2$), the Hessian and Laplace approximation remain equivalent up to a scaling factor $(2r_{\mathrm{MAP}})^{-2}$ via the chain rule.  
Since $c_i$ are bounded, the $O(1)$ scaling in $d,K$ is unaffected.
\end{proof}

\section{Detailed Proof on MMSD}
\subsection{Detailed Proof of Propotion~\ref{prop:mmsd_main_body}}
\label{appendix:mmsd_main_body}
\begin{definition}[MMSD Cross-Sample Mixing Process]
\label{def:mmsd_process}
Let the source data distribution be $\mathbb P_{\mathcal S}(\boldsymbol x, y, e)$, where $\boldsymbol x = (\boldsymbol x_1, \dots, \boldsymbol x_M)$. The MMSD process generates a synthetic feature $\hat{\boldsymbol z}$ and its corresponding origin label $\boldsymbol o$ as follows (a simplified version):
\begin{enumerate}
    \item Draw two samples independently from the source data: $(\boldsymbol x_a, y_a, e_a) \sim \mathbb P_{\mathcal S}$ and $(\boldsymbol x_b, y_b, e_b) \sim \mathbb P_{\mathcal S}$. Note that this draw is cross-sample, and therefore potentially cross-domain and cross-label.
    \item Extract their features using the encoder $\boldsymbol\Phi$: $\boldsymbol z_a = \boldsymbol\Phi(\boldsymbol x_a)$ and $\boldsymbol z_b = \boldsymbol\Phi(\boldsymbol x_b)$.
    \item Construct a synthetic feature $\hat{\boldsymbol z}$ by applying the mixing operator $\mathcal{M}_{\pi, \mathcal{U}, \mathcal{V}}$, which performs cross-frequency mixing, random token sampling, and spatial permutation $\pi$. The origin label $\boldsymbol o$ tracks the source (sample, modality, frequency band, original position) of each token in $\hat{\boldsymbol z}$.
\end{enumerate}
The MMSD training distribution is the distribution of these synthetic pairs, $\mathbb P_{\text{MMSD}}(\hat{\boldsymbol z}, \boldsymbol o)$. The MMSD loss is the expected error in predicting $\boldsymbol o$ from $\hat{\boldsymbol z}$: $\mathcal L_{\text{MMSD}} = \mathbb{E}_{(\hat{\boldsymbol z}, \boldsymbol o)\sim \mathbb P_{\text{MMSD}}} \bigl[ \mathcal{H}(\boldsymbol o, f_{\text{dec}}(\hat{\boldsymbol z})) \bigr]$, where $\mathcal{H}$ is the cross-entropy/L2 loss.
\end{definition}

Our proof hinges on the idea that to succeed at this task, the features must be ``self-contained.'' We formalize this with an assumption:

\begin{assumption}[Feature Sufficiency for Self-Identification]
\label{assum:sufficiency}
An ideal disentangled feature extractor $\boldsymbol\Phi^*$ produces modality-specific features $\phi^*_m(\boldsymbol x_m)$ that are sufficient for self-identification. This means that the information required to identify the origin of a feature's components (e.g., its source modality, frequency band) is contained entirely within the feature itself, without reference to features from other modalities. 
Mathematically, the mutual information between the feature and its origin is maximal, and adding features from other modalities provides no additional information for self-identification:
\begin{equation}
I\Bigl(\phi_m^*(\boldsymbol x_m); \boldsymbol o_m \mid \phi_j^*(\boldsymbol x_j)\Bigr) = I\Bigl(\phi_m^*(\boldsymbol x_m); \boldsymbol o_m\Bigr), \quad \forall j \neq m.
\end{equation}
\end{assumption}

Within this assumption, we can derive the following proposition:

\begin{proposition}[MMSD Minimizes Modal Synergy Risk]
\label{prop:mmsd_minimizes_risk2}
Let the decouplers $f_{\text{dec}}$ be powerful enough to approximate the Bayes-optimal predictor for the auxiliary task. Minimizing the MMSD loss $\mathcal L_{\text{MMSD}}$ with respect to the feature extractor $\boldsymbol\Phi$ is equivalent to minimizing the conditional entropy $\mathbb{E}[\mathcal{H}(\boldsymbol o \mid \hat{\boldsymbol z})]$. This, in turn, drives the learned joint feature distribution $\mathbb P_{\mathcal S}(\boldsymbol\Phi)$ to approximate a factorized distribution, thus minimizing $\mathcal R_{\mathrm{syn}}$.
\end{proposition}

\begin{proof}
The Bayes-optimal decoupler $f_{\text{dec}}^*$ predicts the posterior distribution $P(\boldsymbol o \mid \hat{\boldsymbol z})$. The minimum achievable MMSD loss is the conditional entropy of the origins given the mixed feature, $\mathbb{E}_{\hat{\boldsymbol z}}[\mathcal{H}(P(\boldsymbol o \mid \hat{\boldsymbol z}))]$. To minimize this loss, the encoder $\boldsymbol\Phi$ must produce features that make this posterior as sharp (low-entropy) as possible.

Let's analyze the posterior for a single token's origin, e.g., predicting the modality of its high-frequency component, $o_{\text{high}}$. The mixed token $\hat{\boldsymbol z}_p$ is constructed from independent sources, say $\phi_i(\boldsymbol x_a)$ and $\phi_j(\boldsymbol x_b)$. The posterior is:
\begin{equation}
P(o_{\text{high}}=i \mid \hat{\boldsymbol z}_p) \propto P(\hat{\boldsymbol z}_p \mid o_{\text{high}}=i) P(o_{\text{high}}=i).
\end{equation}
Consider an encoder $\boldsymbol\Phi$ that learns a spurious correlation, meaning $\phi_i$ and $\phi_j$ are statistically dependent on the source domain $\mathbb P_{\mathcal S}$. For such an encoder, the representation $\phi_i(\boldsymbol x_a)$ is not self-contained; its interpretation depends on its correlated counterpart $\phi_j(\boldsymbol x_a)$.

When a mixed feature $\hat{\boldsymbol z}_p$ is created, this correlation is broken because it combines components from independent samples $\boldsymbol x_a$ and $\boldsymbol x_b$. The feature $\phi_j(\boldsymbol x_b)$ provides no information about $\phi_i(\boldsymbol x_a)$. The decoupler, observing a feature component from $\phi_i(\boldsymbol x_a)$ in an alien context provided by $\phi_j(\boldsymbol x_b)$, will be uncertain about its origin. This results in a high-entropy posterior $P(\boldsymbol o \mid \hat{\boldsymbol z})$ and consequently a high $\mathcal L_{\text{MMSD}}$.

Therefore, to minimize the loss, the encoder $\boldsymbol\Phi$ must learn to discard these spurious, context-dependent correlations and produce features that satisfy Assumption~\ref{assum:sufficiency}. The learned representation $\phi_m(\boldsymbol x_m)$ for each modality must be self-contained and identifiable on its own, regardless of the context it is mixed with.

A representation where each component $\phi_m(\boldsymbol x_m)$ is self-contained and does not rely on statistical dependencies with other $\phi_j(\boldsymbol x_j)$ for interpretation is, by definition, one where the joint distribution is factorized. The act of minimizing $\mathcal L_{\text{MMSD}}$ forces the encoder to find a mapping $\boldsymbol\Phi^*$ where the statistical dependence between $\phi_i^*$ and $\phi_j^*$ is uninformative for the prediction task. This directly implies that the mutual information between the feature components is minimized: $I(\phi_i^*; \phi_j^*) \to 0$. This leads to the desired factorization:
\begin{equation}
\mathbb P_{\mathcal S}(\boldsymbol\Phi^*(\boldsymbol x)) \;\to\; \prod_{m\in \mathcal M} \mathbb P_{\mathcal S}(\phi_m^*(\boldsymbol x_m)).
\end{equation}
As the learned joint distribution approaches the product of its marginals, the JS-Divergence term $D_{\text{JS}}$ in $\mathcal R_{\mathrm{syn}}$ decreases towards zero.
\end{proof}

\end{document}